\documentclass{article}
\usepackage[a4paper, total={6in, 9in}]{geometry}

\usepackage[utf8]{inputenc}		
\usepackage[T1]{fontenc}
\usepackage{natbib}
\setcitestyle{authoryear,round,citesep={;},aysep={,},yysep={;}}

\usepackage{graphicx}			
\usepackage{subcaption}
\usepackage{enumerate}
\usepackage{enumitem}
\usepackage[font={small}]{caption}
\usepackage{csquotes}
\usepackage{tcolorbox}
\usepackage{framed}
\usepackage{graphicx}
\usepackage{algorithm}
\usepackage{algpseudocode}
\usepackage{booktabs}
\usepackage{array}

\usepackage{amsmath}			
\usepackage{amssymb}			
\usepackage{amsfonts}			
\usepackage{nicefrac}			

\usepackage[pdftex,pdfborder={0 0 0},
			colorlinks=true,
			linkcolor=blue,
			citecolor=red,
			pagebackref=true,
			]{hyperref}	

\colorlet{leftbarcolor}{black}

\newlength{\leftbarwidth}
\newlength{\leftbarsep}
\renewenvironment{leftbar}{%
    \MakeFramed {\advance \hsize -\width \FrameRestore }%
}{%
    \endMakeFramed
}

\colorlet{leftbarcolor}{blue}

\usepackage{hyperref}       

\usepackage{hyperref}
\hypersetup{
    colorlinks = true,
    linkcolor = red,
    anchorcolor = blue,
    citecolor = blue,
    filecolor = blue,
    urlcolor = blue
    }

\newcommand{\colorremark}{darkgray}
\newcommand\remark[1]{
\setlength{\leftbarsep}{6pt}
\colorlet{leftbarcolor}{\colorremark}
\begin{leftbar}
    {\color{\colorremark}\textbf{Remark: }} #1
\end{leftbar}
}
\newcommand{\colorgoodpractice}{teal}
\newcommand\goodpractice[1]{
\setlength{\leftbarsep}{6pt}
\colorlet{leftbarcolor}{\colorgoodpractice}
\begin{leftbar}
    {\color{\colorgoodpractice}\textbf{Good practice: }} #1
\end{leftbar}
}

\title{Language Evolution with Deep Learning \\
\vspace{0.1em}\small Chapter to appear in the \textit{Oxford Handbook of Approaches to Language Evolution}}

\date{}

\usepackage{authblk}

\author[1]{\textbf{Mathieu Rita}\thanks{Corresponding Author: mathieu.rita@inria.fr}}
\author[2]{\textbf{Paul Michel}}
\author[2]{\textbf{Rahma Chaabouni}}
\author[5]{\textbf{Olivier Pietquin}}
\author[3,4]{\textbf{Emmanuel Dupoux}}
\author[5]{\textbf{Florian Strub}}
\affil[1]{INRIA, Paris}
\affil[2]{Google DeepMind}
\affil[3]{Meta AI Research}
\affil[4]{EHESS,ENS-PSL,CNRS,INRIA}
\affil[5]{Cohere}

\begin{document}


\maketitle

\begin{abstract}

Computational modeling plays an essential role in the study of language emergence. It aims to simulate the conditions and learning processes that could trigger the emergence of a structured language within a simulated controlled environment. Several methods have been used to investigate the origin of our language, including agent-based systems, Bayesian agents, genetic algorithms, and rule-based systems. This chapter explores another class of computational models that have recently revolutionized the field of machine learning: deep learning models. 
The chapter introduces the basic concepts of deep and reinforcement learning methods and summarizes their helpfulness for simulating language emergence. It also discusses the key findings, limitations, and recent attempts to build realistic simulations.
This chapter targets linguists and cognitive scientists seeking an introduction to deep learning as a tool to investigate language evolution.{\begin{center} Supplementary technical materials can be found at \url{https://github.com/MathieuRita/LangageEvolution_with_DeepLearning}
\end{center}}
\end{abstract}

\section{Introduction}
Social animals have been found to use some means of communication to coordinate in various contexts: foraging for food, avoiding predators, mating, etc.~\citep{hauser1996evolution}. Among animals, however, humans seem to be unique in having developed a communication system, natural language, that transcends these basic needs and can represent an infinite variety of new situations~\citep{hauser2002faculty} to the extent that language itself becomes the basis for a new form of evolution: cultural evolution. Understanding the emergence of this unique human ability has always been a vexing scientific problem due to the lack of access to the communication systems of intermediate steps of hominid evolution~\citep{Harnad1976OriginsAE,bickerton2007language}. In the absence of data, a tempting idea has been to reproduce experimentally the process of language emergence in either humans or computational models~\citep{steels1997synthetic,MyersScotton:2002,kirby2002natural}.

\bigbreak

Experimental paradigms with humans~\citep{kirby2008cumulative,raviv2019larger,motamedi2019evolving} have produced significant insights into language evolution. Still, their scope is limited due to the inability to replicate key aspects of language evolution, such as communication within and across large populations and the study of long evolutionary timescales. Computer modeling can help overcome these limitations and has played a prominent role in studying language evolution for a long time~\citep{lieberman1971speech}. In particular, agent-based modeling has been used from the early days of the language evolution research ``renaissance''~\citep{hurford1989biological,steels1995self} and is still a very active and influential field~\citep{reali2009evolution,reali2010words,smith2003complex,vogt2009modeling,gong2014modelling,ke2008language,brace2015achieving,cuskley2017regularity,kirby2015compression}.

\bigbreak

Meanwhile, in the last decade, the field of machine learning has rapidly developed with the advent of deep learning. Deep neural networks have achieved human-level performance in various domains, including image recognition~\citep{he2016deep,chen2020simple}, natural language processing~\citep{devlin2018bert,brown2020language}, automatic translation~\citep{bahdanau2014neural,vaswani2017attention}, and reinforcement learning~\citep{silver2016mastering}.

\bigbreak
This chapter aims to introduce the technical and conceptual background required for using deep learning to simulate language evolution, that is, to simulate both the emergence of communication in evolutionary timescales and patterns of language change in historical timescales~\citep{kottur2017natural,lazaridou2018emergence,lazaridou2020emergent}

First, we present how to implement a communication game (Sec.~\ref{sec:implementing_Lewis_Game}), including formalizing it as a machine learning problem (Sec.~\ref{subsec:comm_games_ML}), designing neural network agents (Sec.~\ref{subsec:design_agents}) and making agents learn to solve the game (Sec.~\ref{subsec:design_optim}).
Second, we examine the Visual Discrimination Game~\citep{lewis1969convention} as a case study (Sec.~\ref{sec:case_study}), which has been widely explored in neural emergent communication research.
Finally, we provide an overview of recent emergent communication simulations with neural networks, highlighting the successes, limitations, and future challenges (Sec.~\ref{sec:research}).

\section{Designing communication games with Deep Learning}
\label{sec:implementing_Lewis_Game}

Communication games~\citep{lewis1969convention, steels1995self, baronchelli2010modeling} are a framework used to investigate how perceptual, interactive, or environmental pressures shape the emergence of structured communication protocols~\citep{kirby2008cumulative,cuskley2017regularity,raviv2019larger}.
This framework has primarily been studied over the past $50$ years and is still one of the leading simulation frameworks in language evolution. See Chapter \textit{Communication games: Modelling language evolution through dyadic interaction} for more details. This section presents how to simulate communication games using Deep Learning. First, we frame the communication game as a multi-agent problem, where each agent is represented by a deep neural network~(Sec.~\ref{subsec:comm_games_ML}). Second, we define communicative agents (Sec.~\ref{subsec:design_agents}). 
Third, we use machine learning optimization to train agents to solve the communication game (Sec.~\ref{subsec:design_optim}).

\subsection{Framing communication games as a machine learning problem}
\label{subsec:comm_games_ML}

\subsubsection{Machine learning is well suited for simulating communication games}
\label{subsubsec:ML_comm}

\begin{figure}[ht!]
    \centering
    \includegraphics[width=0.8\textwidth]{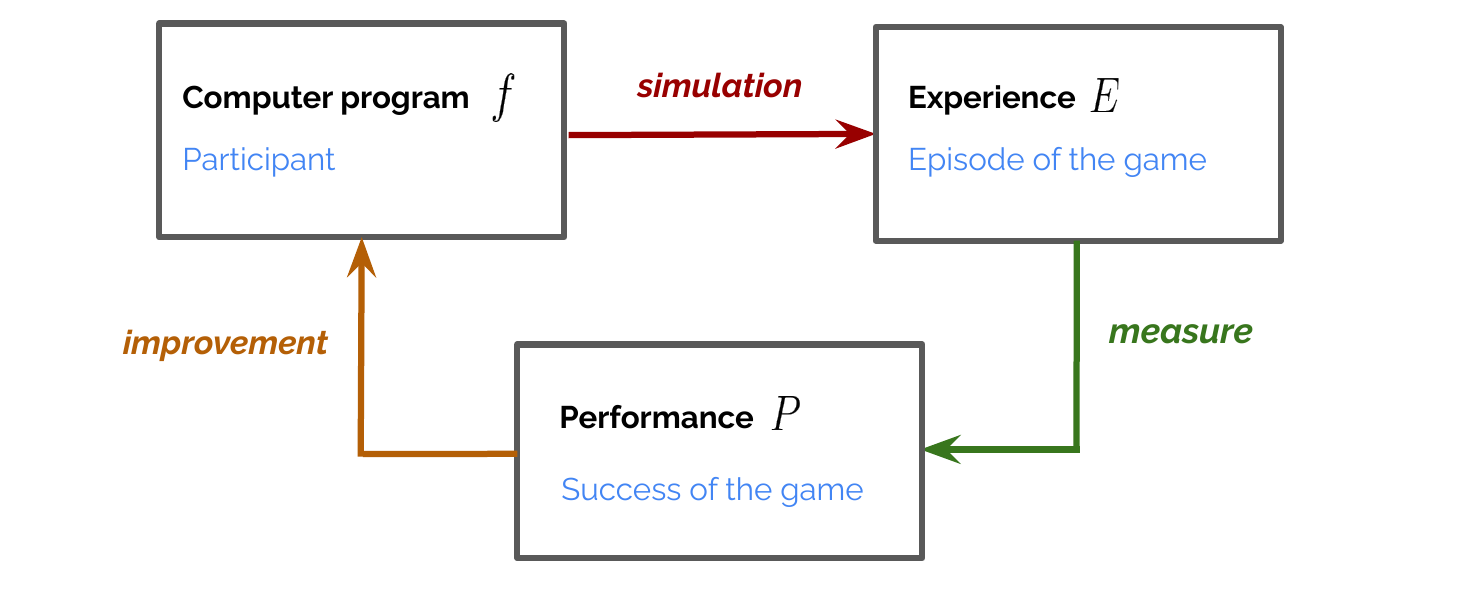}
    \caption{Iterative learning process in a machine learning problem. \textbf{Step 1 (simulation):} The computer program $f$ performs an experience $E$ of the task $T$. \textbf{Step 2 (measure):} The task's success is measured through a performance measure $P$. \textbf{Step 3 (improvement):} Based on its performance, the computer program $f$ update  to improve its future performance, i.e. learns. Communication games can be framed as a machine learning problem by modeling agents as computer programs. The experience $E$ corresponds to an episode of the game, while a performance measure $P$ measures the game's success.}
    \label{fig:iterative_learning_loop}
\end{figure}

\cite{mitchell1997machine} defines machine learning as follows:

\begin{displayquote}
        ``A computer program $f$ is said to learn from an experience $E$ with respect to some class of tasks $T$ and performance measure $P$, if its performance at tasks in $T$, as measured by $P$, improves with experience $E$.''
\end{displayquote}


Machine learning is well suited to frame communication games: participants develop a language through trial and error during a communication game. They iteratively adapt their language production and understanding to achieve a given task for which at least one agent lacks information~\citep{tadelis2013game}. While game theoretic approaches analyze stable communication protocols \citep{crawford1982strategic,skyrms2010signals}, studying the dynamic learning process is a more challenging and richer problem. Borrowing \cite{mitchell1997machine} notations, this dynamic process can be framed as a machine learning problem where participants are computer programs $f$ that perform the communication game $T$. The game's success is measured by $P$ after each episode $E$ of the game, and participants update their communication protocol based on the outcome. After enough iterations, the participants may \textit{converge}, i.e., stabilize on a successful communication protocol, allowing them to solve the game. This iterative learning process is illustrated in Figure~\ref{fig:iterative_learning_loop} and is the fundamental idea of machine learning.

\subsubsection{Formalizing communication games as a machine learning problem}
\label{subsubsec:formal_comm_games}

\begin{figure}[ht!]
    \centering
    \includegraphics[width=0.9\textwidth]{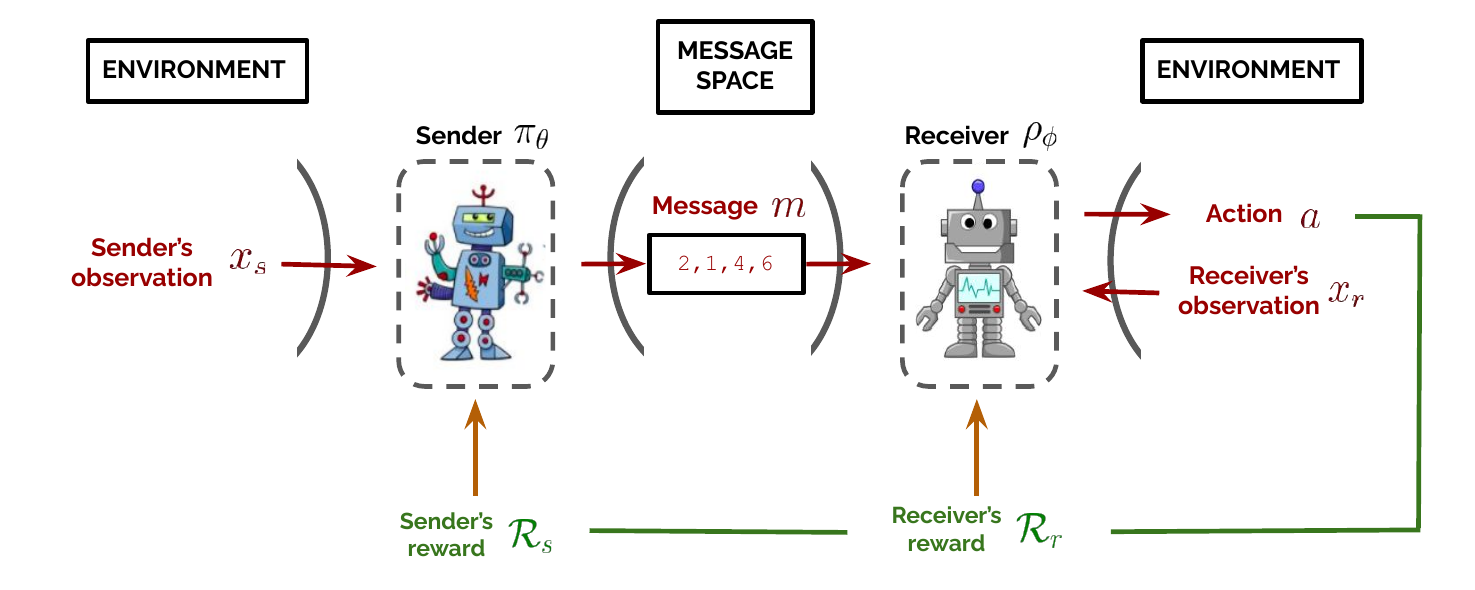}
    \caption{Scheme of a two-player communication game. \textbf{Step 1 (simulation):} Agents play a round of the game: (1) both agents get observations from the environment, (2) The sender sends a message to the receiver, (3) The receiver uses the message and its observation to perform an action in the environment. \textbf{Step 2 (measure):} One reward signal per agent measures the game's success. \textbf{Step 3 (improvement):} Agents receive the reward signals and update their behavior toward better solving the game.}
    \label{fig:comm_game}
\end{figure}

For simplicity, we focus in this chapter on two-player communication games where one agent, the ``sender'' sends messages to a second agent, the ``receiver'' that parses them and takes action to solve the task in an environment\footnote{ This setting is referred to as dyadic unidirectional communication games in the literature~\citep{shannon1948mathematical,harsanyi1967games,kreps,lewis1969convention}.}{.} Formally, the ``sender'' and ``receiver' are parametric models respectively denoted by $\pi_{\theta}$ and $\rho_{\phi}$ with parameters $\theta$ and $\phi$. Both parametric models will further be designed as deep neural networks. As illustrated in Figure~\ref{fig:comm_game}, a round of the game proceeds as follows:
\begin{itemize}
    \item The sender $\pi_{\theta}$ and receiver $\rho_{\phi}$ get observations from their environment denoted by $x_{s}$ and $x_{r}$. 
    \item The sender $\pi_{\theta}$ sends a message $m$ to the receiver $\rho_{\phi}$ where $m$ is a sequence of symbols taken from a fixed vocabulary $\mathcal{V}$.
    \item The receiver $\rho_{\phi}$ uses the message $m$ and its observation $x_{r}$ to perform an action $a$ toward achieving the task.

\end{itemize}

The task's success is then measured by two reward signals $\mathcal{R}_{s}$ and $\mathcal{R}_{r}$ which are given to the sender $\pi_{\theta}$ and the receiver $\rho_{\phi}$ respectively to improve their protocols. Throughout the game, both agents must agree on a common language to solve the game. Importantly, the emergent language is not defined by explicit language rules but implicitly encoded by the sender's parameters $\theta$.

\remark{This chapter presents a simplified formalism of communication games. Rigorously,  communication games should be framed as a special case of Markov Games that provide a broader formal framework for reasoning about multi-agent problems. For further information, refer to \cite{littman1994markov}.}

    \begin{figure}[ht!]
     \centering
     \hfill
     \begin{subfigure}[t]{0.19\textwidth}
         \centering
         \includegraphics[width=\textwidth]{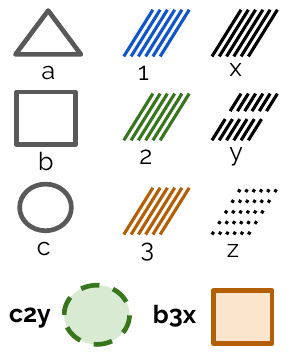}
         \caption{}
         \label{fig:lewis_objects}
     \end{subfigure}
    \hfill
      \begin{subfigure}[t]{0.15\textwidth}
         \centering
         \includegraphics[width=\textwidth]{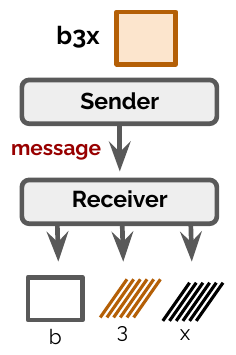}
         \caption{}
         \label{fig:lewis_reconstruction}
     \end{subfigure}
          \hfill
      \begin{subfigure}[t]{0.15\textwidth}
         \centering
         \includegraphics[width=\textwidth]{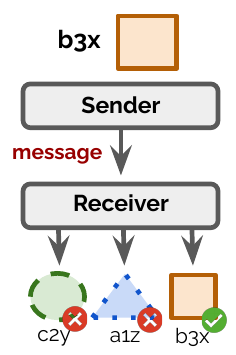}
         \caption{}
         \label{fig:lewis_discrimination}
     \end{subfigure}
     \hfill
    \begin{subfigure}[t]{0.44\textwidth}
         \centering
         \includegraphics[width=\textwidth]{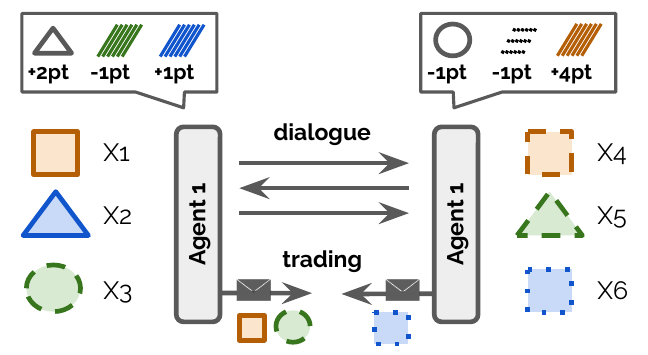}
         \caption{}
         \label{fig:lewis_negociation}
    \hfill
     \end{subfigure}
     \caption{Examples of Lewis and negotiation games. (a) Example of Lewis game object's attributes decomposition (shape, color, style). (b) In the Lewis reconstruction game, the sender observes an object composed of several independent attributes and describes it to the receiver. The receiver must then predict the initial object attributes. (c) In the Lewis discrimation game, the receiver must retrieve the object within a set of distractors. Such a setting does not require manually defining independent attributes, allowing the use of ambiguous real data inputs such as images. (a-c) Such Lewis games usually aim to explore the disentanglement skills of the sender toward producing a compositional language under different scenarios or learning pressures. (d) In negotiation games, agents value objects or attributes differently and get a set of initial objects. They then start dialoguing before executing a final trade. Such tasks involve diverse language interactions such as multi-turn communication, non-fully cooperative games, repeated games, or action binding.}
    \label{fig:game_ex_1}
    \end{figure}

\begin{figure}[ht!]
     \begin{subfigure}[t]{0.31\textwidth}
         \centering
         \includegraphics[width=\textwidth]{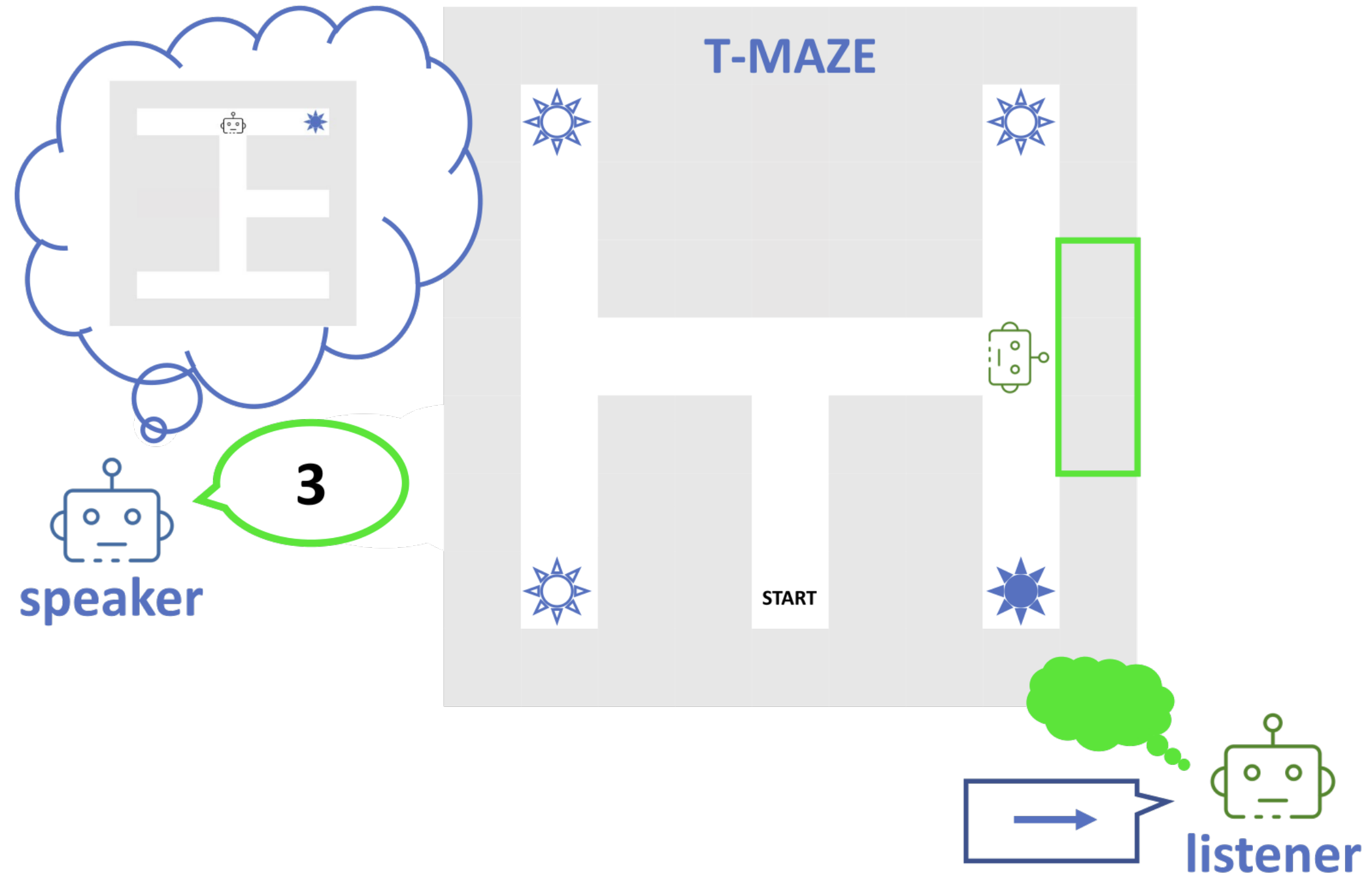}
         \caption{Instruction following games. \\Image from~\citep{kalinowska2022over}}
     \end{subfigure}
          \hfill
     \begin{subfigure}[t]{0.31\textwidth}
         \centering
         \includegraphics[width=\textwidth]{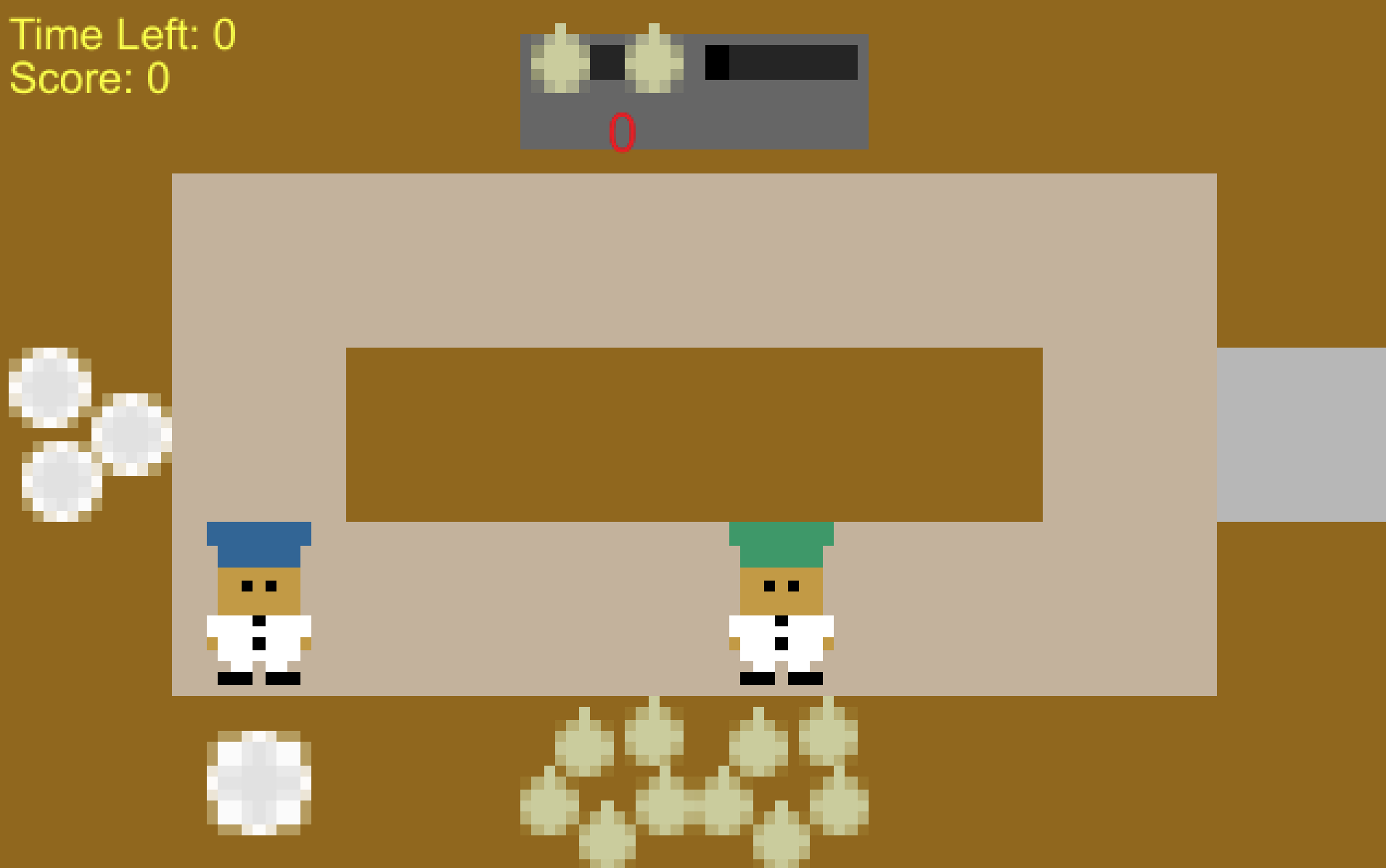}
         \caption{Coordination games. \\Image from~\citep{carroll2019utility}}
     \end{subfigure}
          \hfill
     \begin{subfigure}[t]{0.31\textwidth}
         \centering
         \includegraphics[width=0.5\textwidth]{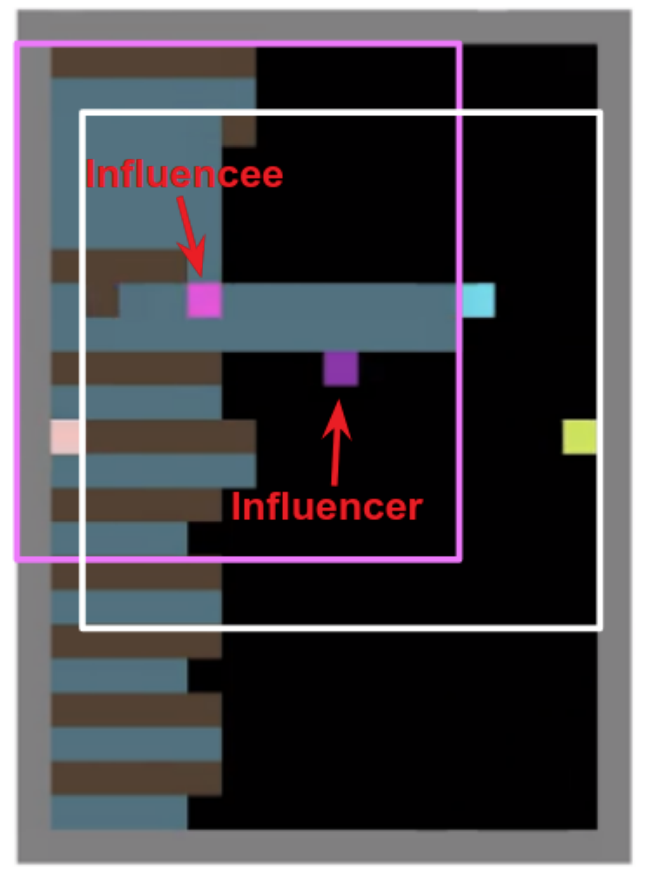}
         \caption{Social dilemma games\\Image from~\citep{jaques2019social}}
     \end{subfigure}
     \caption{Attempts to go beyond Lewis and negotiation games by embodying agents into a 2D world. (a) In the following instruction tasks, the sender is aware of the extensive state of the world and must instruct the receiver on how to reach a predefined goal. Importantly, the receiver only has a partial view of its environment. Such tasks aim to explore how basic embodiment properties may shape communication. (b) In coordination games, the agent needs to communicate to execute joint tasks or improve coordination and success through communication. 
     Such tasks ground language to actions. 
     (c) In social dilemma games, agents are surrounded by multiple agents and must behave accordingly to survive. Such tasks explore multi-channel communication or behavioral communication through actions.}
     \label{fig:game_ex_2}
\end{figure}

In a communication game, the deep neural agents aim to build communication and action policies. This is realized by maximizing their reward. The following is therefore needed: 
\begin{enumerate}
    \item Design the communicative agents as neural networks (Sec. \ref{subsec:design_agents})
    \item Train agents to build a shared communication protocol (Sec. \ref{subsec:design_optim})
\end{enumerate}

Figures~\ref{fig:game_ex_1} and~\ref{fig:game_ex_2} represent communication games commonly studied in language emergence simulations with deep learning. The former presents simple Lewis and negotiation games, while the latter showcases efforts to build more realistic scenarios.

\remark{At the time of writing, many Python libraries, like PyTorch~\citep{paszke2017automatic} and Jax~\citep{jax2018github}, are used for easy implementation and optimization of neural networks and are particularly helpful for beginners due to the abundance of online examples.}

\subsection{Designing communicating agents with neural networks}
\label{subsec:design_agents}

\begin{figure}[ht!]
    \centering
    \includegraphics[width=0.8\textwidth]{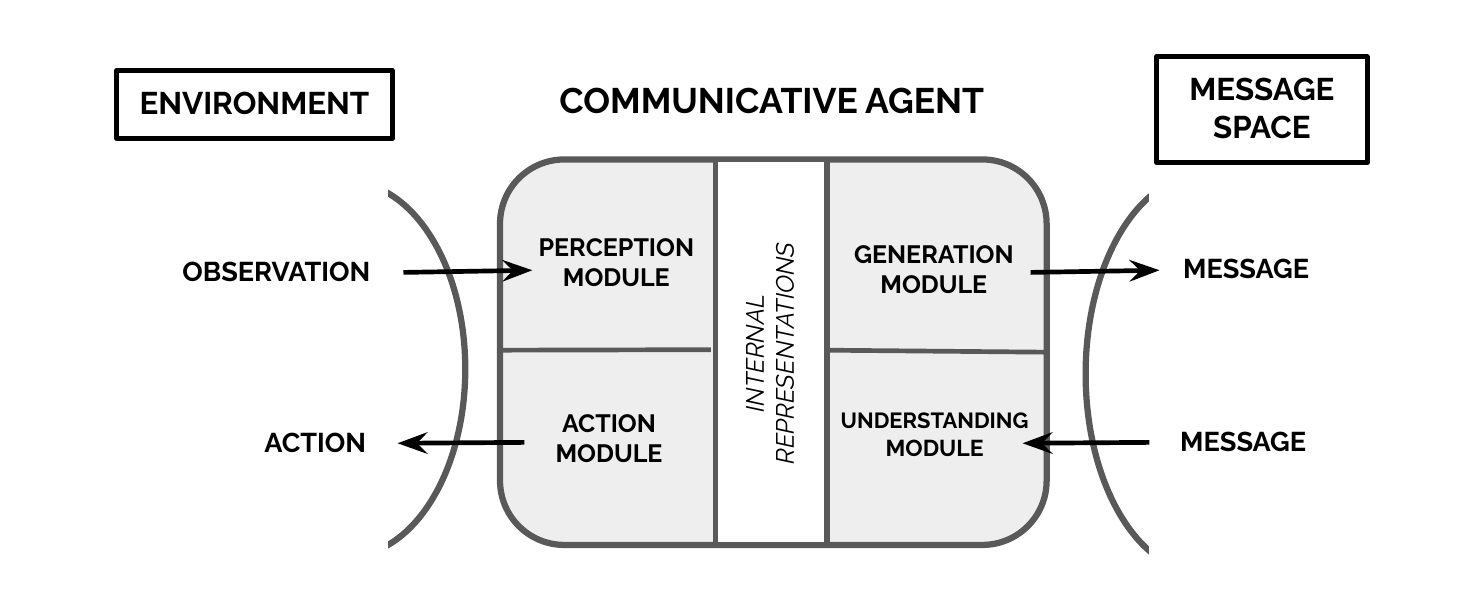}
    \caption{General view of a communicative agent. A communicative agent is composed of four functional modules: a \textbf{perception module} that maps an observation to an \textit{internal representation}; a \textbf{generation module} that maps an \textit{internal representations} to a message; an \textbf{understanding module} that maps a message to an \textit{internal representation}; an \textbf{action module} that maps \textit{internal representations} to an action.}
    \label{fig:designing_agents}
\end{figure}

To model communicative agents, we first break them into functional modules that enable interaction with the environment and other agents (Sec.~\ref{subsubsec:functional_modules}). Then, define neural networks and explain how they can be used to parameterize these functional modules (Sec.~\ref{subsubsec:neural_nets}). Finally, we introduce neural senders and receivers as specific types of neural communicative agents (Sec.~\ref{subsubsec:neural_nets}).

\subsubsection{Designing a communicative agent as functional modules}
\label{subsubsec:functional_modules}

As depicted in Figure~\ref{fig:designing_agents}, a communicative agent should be able to interact with:
\begin{itemize}
    \item \textit{Its environment} by either passively observing it or actively taking actions that influence it ;
    \item \textit{Another agent} using a message space by passively receiving or actively sending messages.
\end{itemize}

Therefore, four functional modules are typically needed to model agents: perception, generation, understanding, and action. (1) The perception module maps an environment's view to an \textit{internal representation}, (2) the generation module generates a message based on internal representations, (3) the understanding module takes a message and builds an \textit{internal message representation}, (4) the action module maps an \textit{internal representation} to an action in the environment. 

Neural networks are suited for modeling and combining these modules.



\subsubsection{Short introduction to neural networks}
\label{subsubsec:neural_nets_intro}

A neural network $f_{\theta}$ is a parametric model approximating a function or probability distribution based on data.
It maps vector inputs to outputs through a succession of linear and non-linear operations. Its learnable parameters $\theta$, called \textit{the weights}, are used to perform the linear operations. The fundamental building block of a neural network is made of two operations:
\begin{itemize}
    \item A \textbf{linear transformation} applying the matrix of weights $\theta^{i} $ to the incoming input:
    \item A \textbf{non-linear transformation} $\sigma$, called the \textit{activation function} (typically sigmoid function, hyperbolic tangent~\citep{lecun1998gradient} or ReLU~\citep{nair2010rectified}): 
\end{itemize}


\begin{figure}[ht!]
    \centering
    \includegraphics[width=0.8\textwidth]{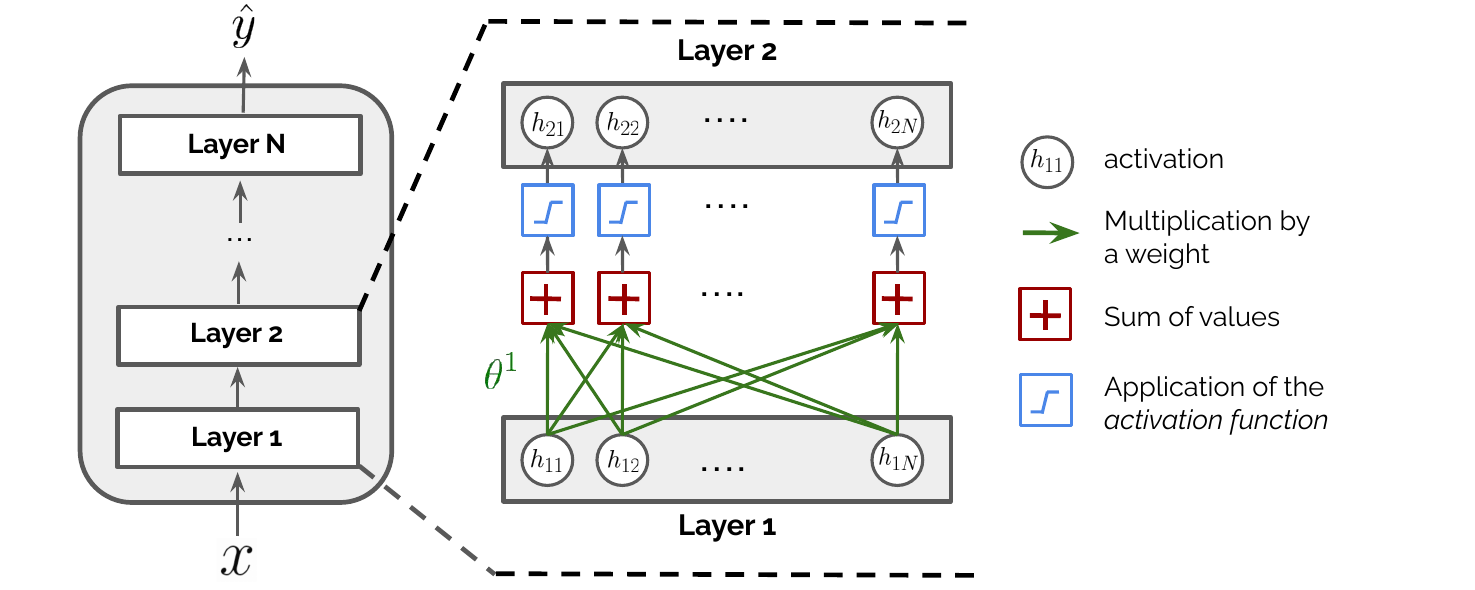}
    \caption{Scheme of a neural network and the operations between two layers. A neural network is a function that takes an input $x=(x_{1},...,x_{n})$ and maps it to an output $\hat{y}$. It comprises several \textit{layers} of \textit{activations}. Each layer results from the application of two operations on the activations of the previous layers: (1) a linear transformation (multiplication by weights and sum), (2) a non-linear transformation (application of a non-linear function). A neural network is parameterized by all the weights acting between each layer.}
    \label{fig:NN}
\end{figure}

As displayed in Figure~\ref{fig:NN}, these operations are stacked at each \textit{layer}, transforming the input $x$ to a prediction $\hat{y}$ through multiple linear and non-linear transformations.

\remark{Neural networks have a crucial property: all operations are differentiable. This allows for using gradient-based methods to learn the weights (see Section~\ref{subsec:design_optim}).
}

\bigbreak

\begin{figure}[ht!]
         \begin{subfigure}[t]{0.48\textwidth}
         \centering
         \includegraphics[width=\textwidth]{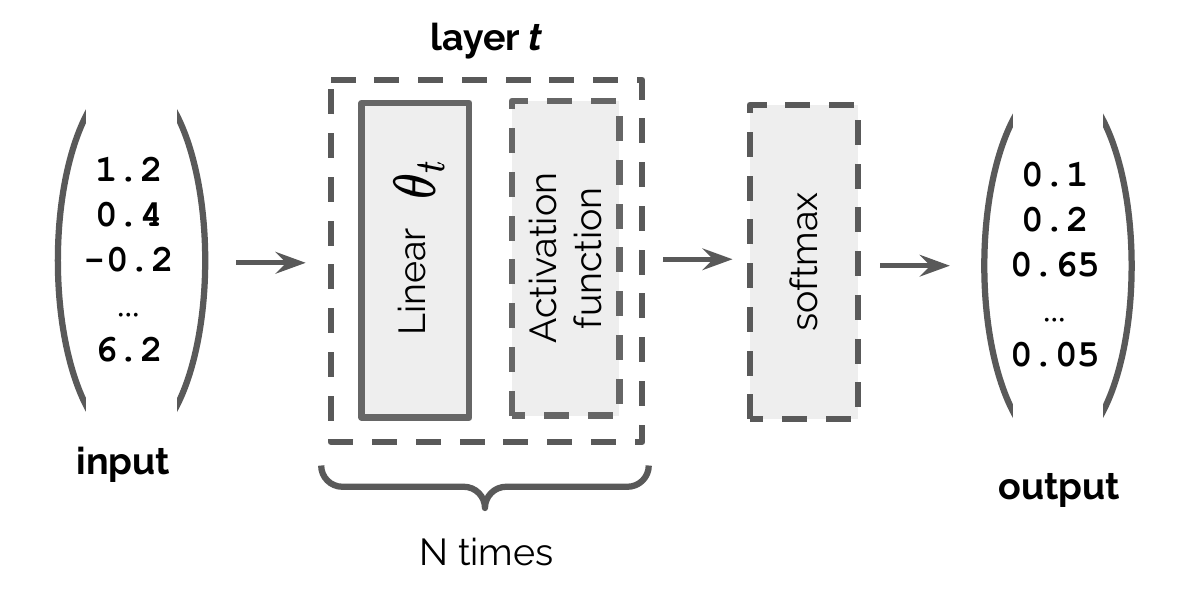}
         \caption{Multi layer perceptron (MLP)}
         \label{fig:MLP}
     \end{subfigure}
     \hfill
     \begin{subfigure}[t]{0.48\textwidth}
         \centering
         \includegraphics[width=\textwidth]{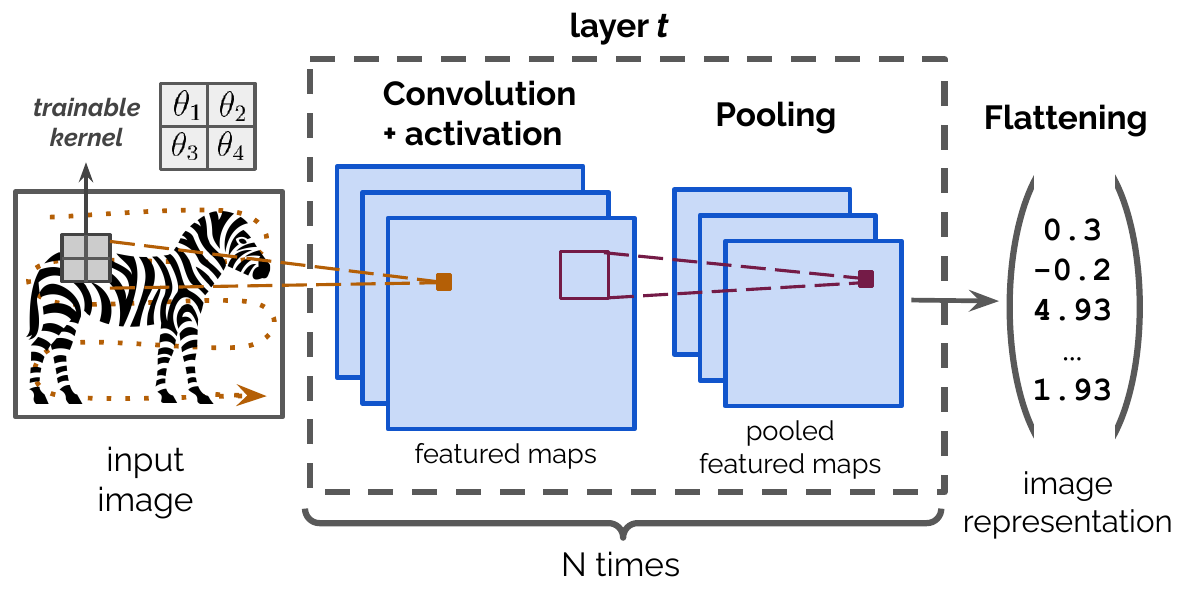}
         \caption{Convolutional Neural Networks (CNN)}
         \label{fig:CNN}
     \end{subfigure}
     \hfill
     \begin{subfigure}[t]{0.48\textwidth}
         \centering
         \includegraphics[width=\textwidth]{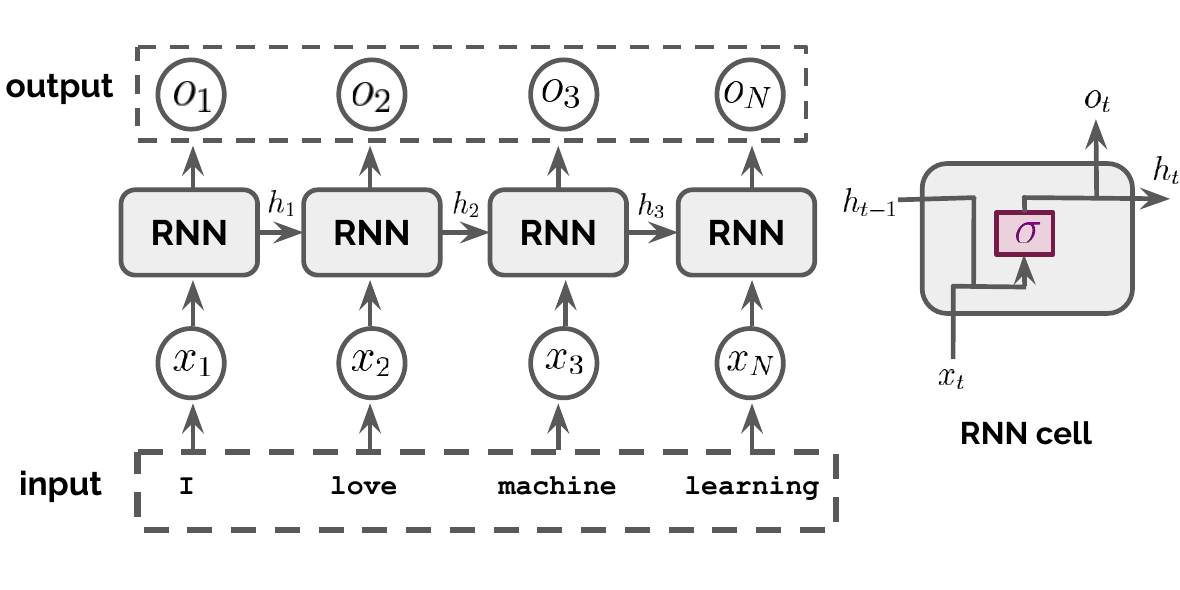}
         \caption{Recurrent Neural Networks (RNN)}
         \label{fig:RNN}
     \end{subfigure}
     \hfill
     \begin{subfigure}[t]{0.48\textwidth}
         \centering
         \includegraphics[width=\textwidth]{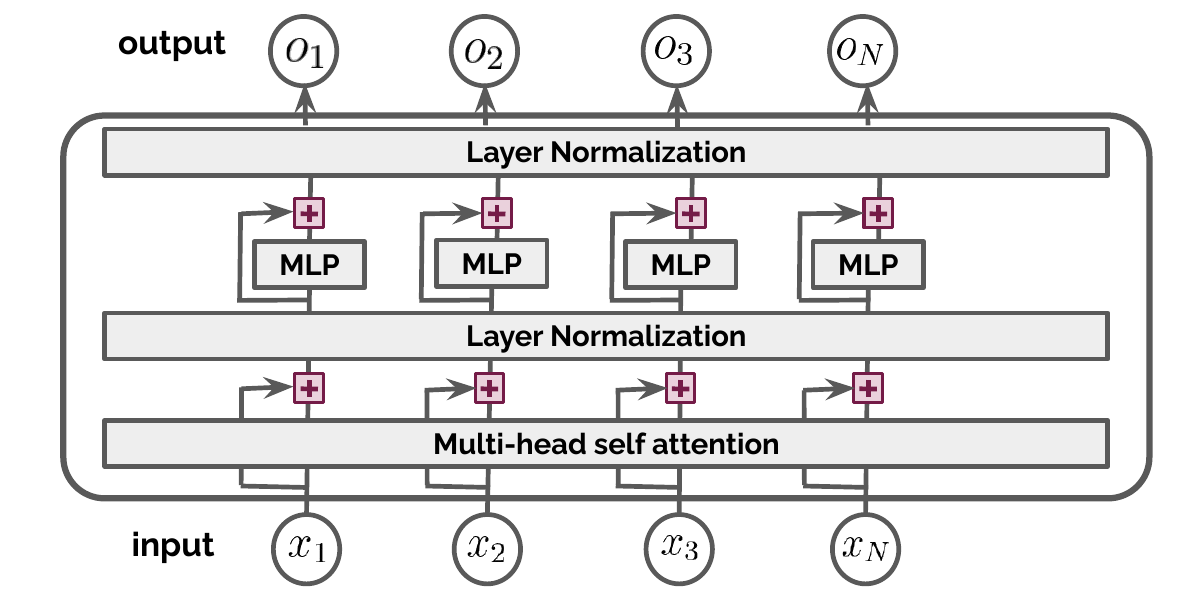}
         \caption{Transformer}
         \label{fig:Transformer}
     \end{subfigure}
     \hfill
     \vskip 1em
     \caption{Classic neural network modules. Other architectures and/or variants such as Graph Neural Networks~\citep{kipf2016semi} for graphs are not presented here.
     \vspace{0.5em}
     \\
     (a) \textbf{Multi-layer perceptron (MLP)}~\citep{rosenblatt1958perceptron} is commonly used to process scalar and heterogeneous input data. It consists of stacked layers, each composed of a linear transformation and a non-linear activation function. A softmax transformation can transform The final activation into a probability distribution.
     \vspace{0.5em}
     \\
     (b) \textbf{Convolutional Neural Network (CNN)}~\citep{lecun1988theoretical} are originally inspired by the visual cortex of animals~\citep{fukushima1980neocognitron}, and are primarily used for visual data. The CNN's core block consists of three operations: a convolution, a non-linear activation function, and an optional downsampling function, aka pooling. The convolution uses a set of filters to scan the input image and detect specific features to build so-called feature maps. The activation function is applied to the feature maps, and then pooling is performed (e.g., taking maximum value over a window of the feature maps). This process is repeated several times, with each subsequent block learning more complex and abstract features. The final layer's output is flattened to provide an abstract input representation. Through the succession of convolutions, the neural network builds a hierarchy of features that capture specific features of the input, e.g., edges and colors. 
     %
     \vspace{0.5em}
     \\
     (c) \textbf{Recurrent Neural Network (RNN)}~\citep{elman1990finding,mikolov2010recurrent,sutskever2011generating} is an architecture designed to process input sequences one element at a time while maintaining an internal state that retains information about the past sequence elements. The memory of the RNN is called a state and denoted by $h_{t}$. It is updated every time the RNN processes an element, allowing it to use information from the past memory $h_{t-1}$. The RNN generates an output representation $o_{t}$ as it processes each new element $x_{t}$ of the ongoing sentence, and the final output $o_{N}$ is often used to represent the entire sequence. However, Vanilla RNNs suffer from training instability when working with long sequences. Alternative architectures with more advanced in-cell operations like LSTMs~\citep{hochreiter1997long} or GRUs~\citep{chung2014empirical}, and regularization methods like layer normalization~\citep{ba2016layer} or gradient clipping~\citep{pascanu2013difficulty} mitigate this issue.
     \vspace{0.5em}
     \\
     (d) \textbf{Transformer}~\citep{vaswani2017attention} is a more recent architecture that processes sequences in parallel rather than sequentially using attention mechanisms. These mechanisms allow selectively focus on different parts of the input sequence when processing it by differently weighting each input element based on its relevance at a given processing step. See \citet{vaswani2017attention} for details on attention.
     Transformers are generally much faster and more efficient than RNNs. Although less intuitive than RNNs, they are replacing recurrent architectures due to their better performances and scalability advantages. Recent dialogue agent successes are based on the Transformer architecture~\citep{brown2020language, hoffmann2022empirical}.
     }
\label{fig:NNs}
\end{figure}

When training a neural network, the goal is to find the optimal weights $\theta$ such that the neural network $f_{\theta}$ accurately maps inputs to their corresponding outputs. Neural networks with enough weights can represent complex functions due to their high expressive power, approximating any continuous function with any level of precision~\citep{hornik1989multilayer}. However, computation or data limitations can hinder this process. Deep learning investigates how to adapt networks' architecture or weight matrix shape to overcome these limitations. Figure~\ref{fig:NNs} presents the main network architectures and the data they are suitable for.

\clearpage

\subsubsection{Neural functional modules}
\label{subsubsec:neural_nets}

Several network architectures can be considered when designing agents modules defined in Sec~\ref{subsubsec:functional_modules}. This section presents some common choices for each module.

\paragraph{Perception module} The perception module maps an observation of the environment to an \textit{internal representation}. The choice of architecture depends on the input observation, which differs across games. For example, a \textit{Convolutional Neural Network}~\citep{lecun1988theoretical} is suitable for generating image representations from visual input data, as illustrated in Figure~\ref{fig:NNs}.


\paragraph{Generation module} The generation module maps an internal representation, i.e., a vector of a given dimension, into a message. 
Recurrent neural networks (RNN)~\citep{elman1990finding,mikolov2010recurrent} and Transformers~\citep{vaswani2017attention} are well suited for sequences and are hence used in standard emergent communication settings~\citep{lazaridou2018emergence,chaabouni2019anti,kottur2017natural,li2019ease,chaabouni2022emergent,rita2022on}. Communication is mainly based on discrete messages, even if some works consider continuous communication protocol~\citep{tieleman2019shaping}.


\remark{To shape the message space, a \textit{vocabulary} of symbols and a \textit{maximum lenght} must be introduced. It's also possible to add an end-of-sentence token \emph{EoS} to indicate the end of the message. When making these design choices, task complexity should be considered; a larger vocabulary and message length allow for communicating more information/concepts, while a smaller vocabulary and message length require 
better information compression and, hence, a more structured communication protocol.}

\paragraph{Understanding module} The understanding module maps a message to an internal representation. Since messages are discrete sequences, RNNs, and Transformers are well-suited for this module.


\paragraph{Action module} The action module maps an internal representation of an action in the environment. Since the internal representations are scalars and actions a finite set of possibilities, a well-suited architecture is the Multi-Layer Perceptron followed by a softmax that draws a probability distribution over the potential actions.

\remark{Deep learning techniques allow training a system composed of multiple differentiable modules end-to-end. The agent is seen as a single block that provides a prediction given input and output data instead of past methods that glue independently trained/designed blocks together. In communication games, the sender and receiver are both fully-differentiable individually. However, the message generation between them does not necessitate \emph{on purpose} to separate the training of the agents. 
Nonetheless, the message generation can still be made differentiable as described in Section~\ref{subsubsec:opt_lewis}.}

\goodpractice{Exploring various neural architectures is a common reflex when starting with deep learning. However, its impact is limited in practice compared to other experimental choices such as task definition, optimization, data, and training objective. Basic neural architectures are recommended to avoid compounding factors when comparing methods.}


\subsubsection{Modeling neural network communicative agents in communication games}
\label{subsubsec:sender_receiver}

\begin{figure}[ht!]
    \centering
    \includegraphics[width=1.\textwidth]{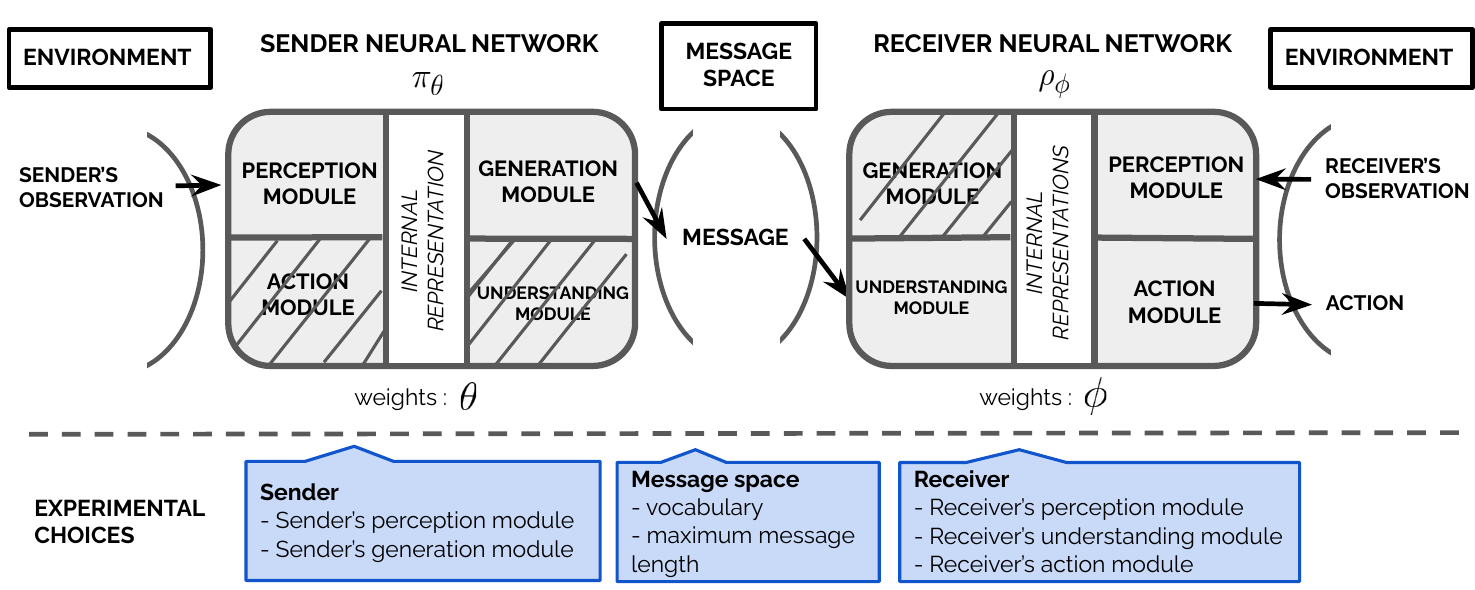}
    \caption{Summary of neural communicative agents modeling in a unidirectional communication game. The sender only uses a perception module to process observations and a generation module to create messages. The receiver uses a perception module to process observations, an understanding module to process messages, and an action module to interact with the environment. A vocabulary and maximum message length are defined for the sender's generation. All the experimental choices are framed in blue.}
    \label{fig:summary_NN}
\end{figure}

Section~\ref{subsec:design_agents} presents the components of a general communicative agent, though not all modules may be used during a game. Figure~\ref{fig:summary_NN} illustrates sender and receiver modeling in a unidirectional game. This modeling is used in the use case we derive in Section~\ref{sec:case_study}, namely the Visual Discrimination Game.

\subsection{Optimizing the agents to solve the game}
\label{subsec:design_optim}

In Deep Learning, the goal is to train neural networks to solve a task, i.e., find the optimal weights that maximize their performance. This section covers optimization techniques for training neural networks and their application to communication games.

\subsubsection{Optimizing a machine learning problem}

\paragraph{Data and learning techniques} To train neural networks, suitable learning techniques must be chosen depending on the task and the availability of training \textit{data}, which consists of input-output pairs $(x,y)$. Two standard techniques used to solve communication games  are: 


\begin{itemize}
    \item \textbf{Supervised Learning (SL)}: The neural network is given a training set $\mathcal{D}_{train} = (x_{n},y_{n})_{n=1}^{N}$ of $N$ input-output pairs $(x_{n},y_{n})$, and its objective is to learn how to map inputs to their corresponding outputs. 
    An example of a supervised language task is the translation: the network learns to map one language to another by training on pairs $x_{n}$ and $y_{n}$, where each pair consists of aligned source and target sentences.
    Supervised learning finds the weights that enable the network to generalize this mapping to new, unseen examples drawn from the same distribution as the training data, e.g., trying to translate beyond the initial corpus. In communication games, Supervised learning tasks often involve classification (e.g., object selection, attribute reconstruction, translation)
    and regression (e.g., drawing, pixel reconstruction).

    \item \textbf{Reinforcement learning (RL)}: In RL, a neural network, or agent, must perform a sequence of actions to resolve a task within its environment. 
    These actions yield \textit{rewards} that gauge the effectiveness of the network's task performance. 
    The network is then optimized to maximize its expected reward, i.e., performing the sequence of actions that lead to the highest task success. 
    Noteworthy, the probability of action is called a policy in RL. 
    In communication games, the sender produces a sequence of symbols to assist the receiver in completing a predetermined task. If this sequence leads to a successful outcome, the sender is rewarded positively; otherwise, it receives a negative reward. Through iterative trial and error, the sender refines its sequence of symbols toward maximizing its reward and ultimately solving the game, as further detailed in Section~\ref{subsubsec:opt_lewis}
\end{itemize}

Supervised learning is easy to apply and highly reproducible but requires a known target. On the other hand, reinforcement learning is more generic and only requires a score to be defined at the cost of being more complex. 
For instance, to train a network to play chess, supervised learning would involve imitating the moves of a pro-player with a dataset~\citep{silver2016mastering}, while reinforcement learning would require playing the whole game and rewarding victories: the training is more complex and slower, but it does not require data~\citep{silver2017mastering}. It is noteworthy that the reinforcement learning reward can be defined arbitrarily, e.g., one may give an extra bonus when winning the game while preserving the queen, or it could also be used on top of a supervised training regime.
This approach has been applied to train large dialogue systems~\citep{ouyang2022training} by imitating the human language and refining it with reinforcement learning.

\paragraph{The loss function} Regardless of the learning technique, the task's success is optimized by introducing a proxy, the \textit{loss function} $l(f_{\theta};x,y)$. The goal is then to find weights $\theta$ such that the neural network $f_{\theta}$ minimizes the average loss function $\mathcal{L}_{\theta}$ over the entire training dataset $\mathcal{D}_{train}$:

\begin{align}
    \min_{\theta} \mathcal{L}_{\theta}, \quad \quad \quad \quad \text{where: }\mathcal{L}_{\theta}=\mathbb{E}_{(x,y)\sim\mathcal{D}_{train}}[l(f_{\theta};x,y)]
    \label{eq:loss_pb}
\end{align}

Loss functions vary depending on the network output and the training task~\citep{bishop2006pattern,goodfellow2016deep}.
In supervised classification tasks, the Cross-Entropy loss is commonly used to measure the difference between the predicted class probabilities and the true class labels. For supervised regression tasks, the Mean Squared Error loss is typically employed to measure the difference between predicted and true values. In reinforcement learning, the losses often include the TD error or the score function~\citep{sutton2018reinforcement}, which converts the expected sum of rewards as a training objective.
%
%
In communication games, we often use either a cross-entropy error for the listener or the score function for the speaker. For instance, the cross-entropy would quantify the error of selecting the wrong object in a referential game. In contrast, the score function would quantify how the speaker policy, i.e., emergent language, should be modified according to the collected rewards to solve the task. We explain further these intuitions in Section~\ref{subsubsec:opt_lewis}.

\paragraph{Optimizing the loss function}

The loss function is reduced using a learning process that involves a series of updates known as \textit{Gradient Descent} updates~\citep{rumelhart1986learning}. They iteratively adjust the network's parameters by following the loss gradient. The magnitude of the update is controlled by a hyperparameter $\eta$ called the \textit{learning rate}. Given the optimization problem \ref{eq:loss_pb}, the goal is to find weights such that the loss gradient equals $0$. This is achieved by repeating the following gradient update rule:

\begin{align}
    \theta_{n+1} = \theta_{n} - \eta \nabla_{\theta}\mathcal{L}_{\theta}|_{\theta_{n}}
\end{align}

where $\theta_{n}$ and $\theta_{n+1}$ are the model parameters respectively at iteration $n$ and $n+1$, $\nabla_{\theta}\mathcal{L}_{\theta}$ the gradient of the loss function $\mathcal{L}_{\theta}$ and $\eta$ the learning rate.

\begin{figure}[ht!]
    \centering
    \includegraphics[width=0.95\textwidth]{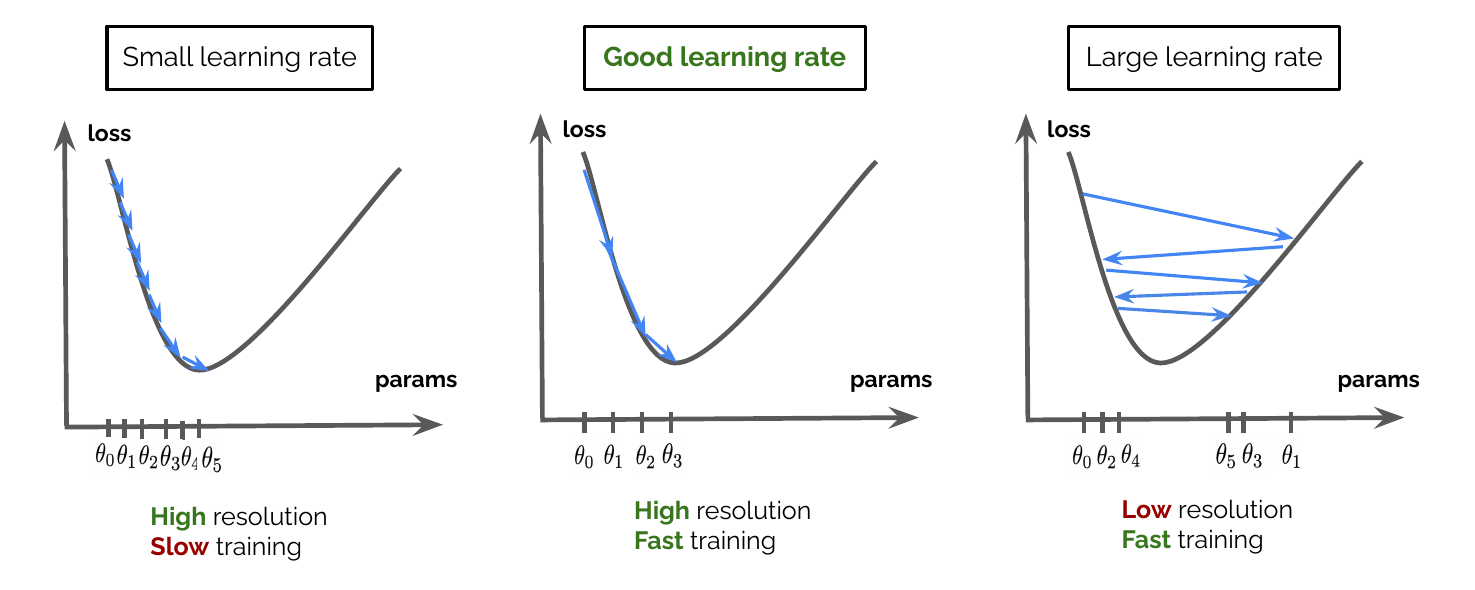}
    \caption{1D representation of the effect of adjusting the learning rate. (a) Small learning rate: parameter updates are too parsimonious; the optimum may be attained, but it requires a large number of updates; (b) Adequate learning rate:  the optimum may be reached with precision while limiting the number of updates; (c) Large learning rate: the training may be initially faster but lack the precision to reach the minimum leading to an oscillatory, and sometimes destructive, effect.}
    \label{fig:LR}
\end{figure}

In practice, computing the exact gradient of the averaged loss function $\nabla_{\theta}\mathcal{L}_{\theta}$ is infeasible since it necessitates processing the complete dataset. \textit{Stochastic Gradient Descent}~\citep{bottou2010large} overcomes this challenge by approximating the loss function gradient using a limited number of data samples, or \textit{batches} at each iteration. In standard machine learning libraries~\citep{jax2018github,paszke2019pytorch}, \textit{Stochastic Gradient Descent} updates are performed by pre-implemented methods referred to as \textit{optimizers}. In communication games, this gradient is the mathematical operation that modifies the agent behavior. For instance, every single speaker update alters its generation of symbols, refining its emergent language step after step toward maximizing the reward objective.

\goodpractice{\textit{What and how to choose training parameters ?}
\begin{enumerate}
    \item \textbf{Choice of the learning rate} The learning rate controls how strongly a model is adjusted given the loss gradient. As illustrated in Figure~\ref{fig:LR}, a too-small learning rate may induce ineffective learning, and a too-large learning rate may cause counter-productive (or even detrimental)~\citep{goodfellow2016deep} updates. Fortunately, adaptive optimizer algorithms are designed to tune the learning rate during training. Typically, the learning rate starts large to kickstart the training and is gradually reduced toward the end for fine-tuning. An effective generic choice is to use optimizer \textit{Adam}~\citep{kingma2014adam} optimizer with an initial learning rate of $4.10^{-3}$. 
    \item \textbf{Choice of batch size} The batch size controls the gradient update accuracy. 
    Counter-intuitively, a large batch size may not guarantee optimal performance.
    In supervised training, a batch size of $64$ to $512$ samples is often recommended, while in reinforcement learning, larger batch sizes of $256$ to $4096$ samples are preferred.
    \item \textbf{Balancing batch sizes and learning rates} The batch size and learning rate values are correlated, with no straightforward recipe for interleaving them. It is often worth jointly sweeping over those two hyperparameters to boost performance.
\end{enumerate}
}

\begin{figure}[ht!]
\centering
    \begin{subfigure}[t]{0.33\textwidth}
         \centering
         \includegraphics[width=\textwidth]{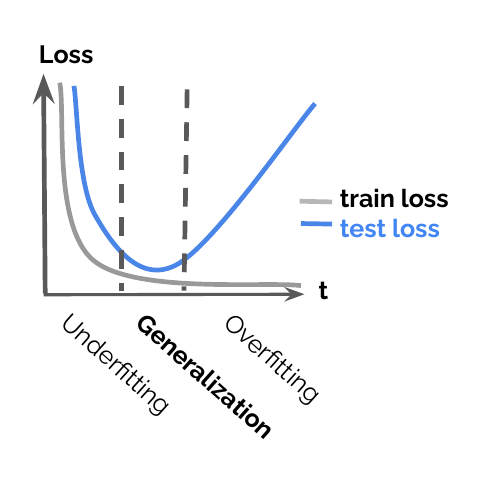}
         \caption{Loss evolution}
         \label{fig:loss_evolution}
     \end{subfigure}
     \hfill
     \begin{subfigure}[t]{0.66\textwidth}
         \centering
         \includegraphics[width=\textwidth]{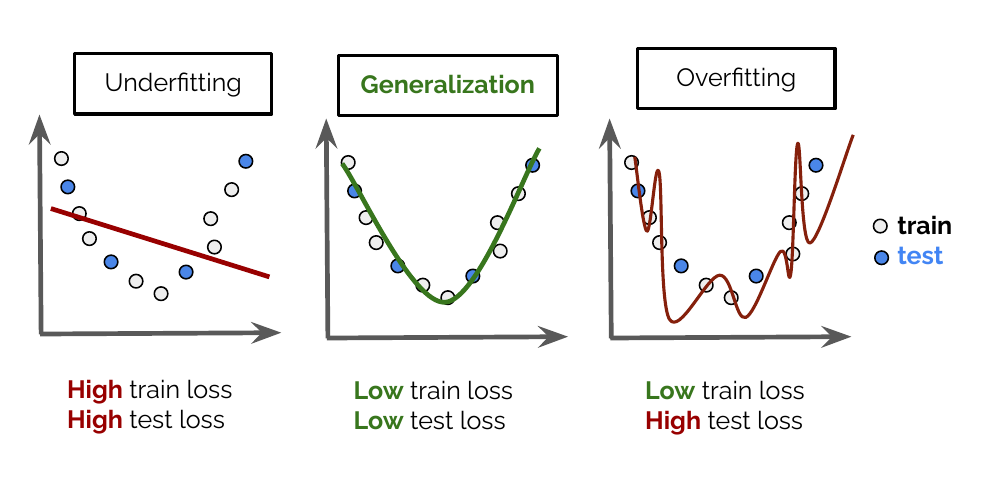}
         \caption{Corresponding fitting curves in the three regimes}
         \label{}
     \end{subfigure}
     \hfill

    \caption{Training regimes for a 1D regression problem trained with the Mean-Squared Error Loss. During training, there are three phases: \textbf{underfitting}: both the train and test loss are high, and the corresponding fitting curve does not match nor the training data nor the test data;  \textbf{generalization}: both the train and test loss are low, and the fitting curve matches both the training and test data;  \textbf{overfitting}: the training loss remains low, but the test loss increases and the corresponding fitting curve perfectly matches the training data but does not generalize to the test data.}
    \label{fig:overfitting}
\end{figure}

\paragraph{Generalization and overfitting} Training a model involves minimizing the loss of the training data, but evaluating its performance on unseen data is crucial to ensure the network's quality. 
Intuitively, it is like creating an exam for students with unseen exercises to ensure they correctly understand the lecture.
ML Practitioners distinguish (1) the training dataset $\mathcal{D}_{train}$ and its corresponding loss $\mathcal{L}_{train}$, (2) the test dataset $\mathcal{D}_{test}$ with unseen samples and its corresponding loss $\mathcal{L}_{test}$. The relation between the two losses indicates how well the model generalizes and can be trusted. Figure~\ref{fig:overfitting} illustrates the three regimes that may occur when training a model:

\begin{itemize}
    \item \textbf{Underfitting}: Both $\mathcal{L}_{train}$ and $\mathcal{L}_{test}$ are high, indicating ineffective learning. An under-parametrized network or a small learning rate may cause persistent under-fitting. In communication games, this scenario arises when no successful communication emerges between the sender and receiver, resulting in a poor task success both on $\mathcal{D}_{train}$ and $\mathcal{D}_{test}$.
    
    \item \textbf{Generalization}: Both $\mathcal{L}_{train}$ and $\mathcal{L}_{test}$ are low, indicating successful training and generalization. In communication games, this regime occurs when agents develop a
    successful communication on $\mathcal{D}_{train}$ that generalizes well to an unseen dataset $\mathcal{D}_{test}$, resulting in high task success both on $\mathcal{D}_{train}$ and $\mathcal{D}_{test}$.
    
    \item \textbf{Overfitting}: $\mathcal{L}_{train}$ is low, but $\mathcal{L}_{test}$ is high, indicating that the network has recorded the training data and is not able to generalize well to new data. This can be addressed by increasing the amount of training data or using regularization techniques, as explained below. In communication games, this regime occurs when agents develop effective communication on $\mathcal{D}_{train}$ but fail to generalize to an unseen dataset $\mathcal{D}_{test}$. 
\end{itemize}

In communication games, the underfitting regime occurs when the emergent language is not powerful enough to resolve the task, i.e., similar sequences of symbols may represent completely different concepts. Conversely, overfitting occurs when a unique sequence of symbols defines each concept without any structure or compositionally. Therefore, there is no generalization beyond the concepts observed at training time. Optimally, we expect the emergent language to generalize to unseen concepts, which may result from emerging compositionality ~\citep{kirby2001spontaneous,rita2022emergent}. 


\paragraph{Monitoring training} When training a model, it is recommended to divide the dataset into three parts: $\mathcal{D}_{train}$, $\mathcal{D}_{val}$, $\mathcal{D}_{test}$ (typical proportion $80/10/10$). $\mathcal{D}_{train}$ is used to train the model, $\mathcal{D}_{val}$ to find the generalization regime, tune hyperparameters, and retrieve the best model across training, $\mathcal{D}_{test}$ is used to test the model and report the final score. Intuitively, validation data is similar to mock exams, whereas test data is the actual network exam. In practice, the validation loss is regularly plotted and when it starts increasing, training is stopped~\citep{bishop2006pattern}. This technique is known as \textit{Early stopping}. 
\paragraph{Regularization methods} 
Regularization methods were developed to prevent potential overfitting~\citep{goodfellow2016deep}, as the number of network parameters can be much larger than the data. Some of the most common techniques include:

\begin{itemize}
    \item \textbf{Weight decay}: Overfitting may be caused by excessively increasing parameters. A weight decay penalty can be applied to the training loss. Using the AdamW variant of the Adam optimizer is recommended to ensure proper integration of the weight penalty~\citep{loshchilov2017decoupled}.
    \item \textbf{Clipping}: Overfitting may be caused by destructive updates due to unexpected large loss gradients. Clipping methods are applied to cope with such events~\citep{pascanu2013difficulty}. 
    \item \textbf{Dropout}: Overfitting may be alleviated by only training subsections of networks for each update. This masking mechanism may be applied at the neuron level~\citep{srivastava2014dropout} or neural-block level for deep networks~\citep{ghiasi2018dropblock}.
    \item \textbf{Normalization layers}: High neural activation inside the network tends to deteriorate the training process and favor overfitting. Normalization layers were developed to recalibrate the neural activations, such as batch-normalization~\citep{ioffe2015batch}, which is a parametrized whitening layer, or layer-normalization~\citep{ba2016layer}. 
    \item \textbf{Data augmentation}: As overfitting often spurs with the lack of data, a common practice is to artificially augment the training set by applying random transformations such as resizing, color alteration, or partial masking for image data~\citep{ba2016layer}.
\end{itemize}

\remark{Applying all regularization techniques simultaneously may seem appealing but can lead to conflicts. For instance, batch normalization should not be applied with dropout, weight decay should not be applied to batch normalization and layer normalization parameters, and strong data augmentations may impair generalization. There is no single rule: finding the right balance for regularization often requires trial and error.}

\begin{figure}[ht!]
    \centering
    \includegraphics[width=0.95\textwidth]{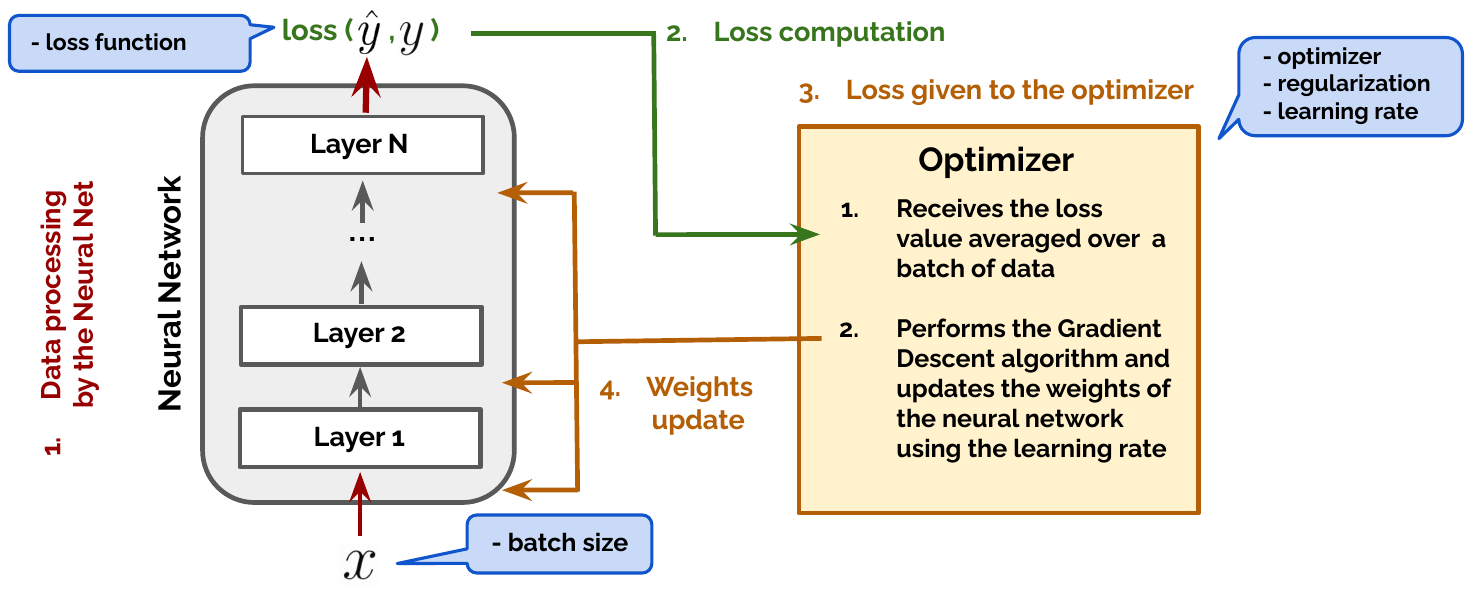}
    \caption{Updates cycle of a neural network in supervised learning. First, the neural network receives a batch of input $x$. The neural network then processes the data and outputs a prediction $\hat{y}$. Second, the loss function computes the average loss value by comparing $\hat{y}$ to the ground truth output $y$. Third, the average loss value is fed to the optimizer. Last, the optimizer performs a Gradient Descent step to update the neural network weights based on the learning rate.  All the experimental choices are framed in blue.}
    \label{fig:optim_pip}
\end{figure}

\paragraph{Summary}: Figure~\ref{fig:optim_pip} summarizes the training process and the associated experimental choices.

\subsubsection{Optimizing communication games with machine learning}
\label{subsubsec:opt_lewis}

Unlike a single network training, two networks are trained simultaneously during a communication game, sometimes requiring different learning methods for each agent.
The process involves selecting appropriate (1) learning methods, (2) rewards and loss functions, and (3) optimization protocols.
%

\remark{The machine learning community has developed frameworks for simulating various communication games, which can be rapidly replicated, understood, and modified. Existing codebases include~\citep{kharitonov2019egg} and ~\citep{chaabouni2022emergent} as long as the detailed notebook we provide.}

\paragraph{Learning methods} Three learning pipelines are mainly used to train agents in communication games:
\begin{enumerate}
    \item \textbf{Both agents optimized with RL}: This generic and realistic setting assumes no specific task format and involves separate agents with individual rewards and training losses, making it suitable for training any task. However, such training is usually hard to optimize with high variance and requires careful use of RL tools we introduce later. 
    \item \textbf{Sender optimized with RL and Receiver optimized with SL}: This approach is well-suited for single-turn message games where the receiver only needs to perform one valid action after receiving a message, such as in referential games~\citep{lewis1969convention,skyrms2010signals}. In such cases, the receiver's action $a$ is fully determined by the sender's observation $x_{s}$ and its message $m$, creating a supervised training sample $(m,a)$ for the receiver. The receiver's training becomes more robust by learning to map messages $m$ to the corresponding output actions $a$ using a supervised loss. Note that the sender still needs to be optimized with RL since message generation is non-differentiable, i.e., the receiver's error cannot propagate to the sender. It ensures more stable training than using a pure RL reward-based approach. 
    \item \textbf{Both agents optimized with SL}: When both agents cooperate fully and optimize the same learning signal, they can be trained using a single supervised training signal. In this scenario, the Sender-Receiver couple is optimized as a single network that maps inputs $x_{s}$ to output actions $a$, with a discrete intermediate layer. Reparametrization tricks such as Gumbel-Softmax~\citep{jang2016categorical,maddison2016concrete} have been developed to overcome the non-differentiability of message generation and allow the receiver's error to flow to the sender\footnote{This is the same approach as training a Variational Auto-Encoder (VAE)~\citep{kingma2013auto}}{.} Although this approach is more stable than RL methods, we won't go into details because it assumes a less realistic training hypothesis,  e.g., the exact error is propagated between sender and receiver as if they were mentally connected.
 
\end{enumerate}

We next derive the case where agents are optimized with RL as it covers all communication tasks.

\paragraph{Reward} 
Reward functions $\mathcal{R}_{s}$ and $\mathcal{R}_{r}$ must be defined to measure the success of the communication task for each agent. 
These functions typically take agents' observations $x_{s}$ and $x_{r}$ and the receiver's action $a$ as input and return $1$ if the task is solved, $0$ otherwise.

\remark{The reward is the core element inducing the structure of the emergent language. Thus, we recommend carefully avoiding designing rewards toward obtaining a specific language, e.g., directly rewarding compositionality or syntactic properties. Instead, we suggest using rewards that measure communication success without any human prior. Hence, language features may emerge from solving a specific task rather than being forced by design. 
}

The agents' goal is to maximize their respective reward over time, i.e., the expected rewards:
\begin{align}
        \left\{
    \begin{array}{lll}
        \mathbb{E}_{\tau \sim \textrm{game}(\pi_{\theta},\rho_{\phi})}[\mathcal{R}_{s}] & \text{Sender's expected reward}  \\
        \mathbb{E}_{\tau \sim \textrm{game}(\pi_{\theta},\rho_{\phi})}[\mathcal{R}_{r}] & \text{Receiver's expected reward} 
    \end{array}
\right.
\end{align}

$\tau \sim \textrm{game}(\pi_{\theta},\rho_{\phi})$ denotes a game episode that depends on the sender's and receiver's stochastic policies.
The sender message $m$ and the receiver's action $a$ are sampled from those distributions.

\remark{
\begin{itemize}
    \item \textbf{Complete formalism:} A complete writing of the expectations should be:
    \begin{align}
        \left\{
    \begin{array}{lll}
        \mathbb{E}_{x_{s} \sim o_{s},x_{r} \sim o_{r},m\sim \pi_{\theta}(.|x_{s}),a \sim \rho_{\phi}(.|m,x_{r})}[\mathcal{R}_{s}(x_{s},x_{r},a)] \\
        \mathbb{E}_{x_{s} \sim o_{s},x_{r} \sim o_{r},m\sim \pi_{\theta}(.|x_{s}),a \sim \rho_{\phi}(.|m,x_{r})}[\mathcal{R}_{r}(x_{s},x_{r},a)]
    \end{array}
\right.
    \end{align}

where the game is instantiated by sampling the initial agent observations $x_{s}$ and $x_{r}$, the message $m$ is sampled according to the sender's policy $\pi_{\theta}(.|x_{s})$ and the receiver's action $a$ is sampled according to its policy $\rho_{\phi}(.|m,x_{r})$. 
\item \textbf{Stochasticity:} Policy stochasticity is crucial in RL training as it enables the agent to explore its message/action space and learn from its errors. This implies that an object can be described by several messages, and a receiver may take different actions based on a given message.
\end{itemize}
}

%

\paragraph{Loss functions \& gradient updates}

In reinforcement learning, the goal is to minimize the expected negative reward. However, this objective cannot be directly turned into a gradient update as the reward is not differentiable by definition. 
Mathematical tools have been developed to circumvent this issue~\citep{sutton2018reinforcement}. 
The policy-gradient algorithm~\citep{sutton1999policy} is mostly used in neural language emergence. Denoting by $\nabla_{\theta}\mathcal{L}_{\theta}$ and $\nabla_{\phi}\mathcal{L}_{\phi}$ the sender and receiver's respective loss gradient, we have:
    \begin{align}
        \left\{
    \begin{array}{lll}
        \nabla_{\theta}\mathcal{L}_{\theta} & = - \mathbb{E}_{\tau}[\nabla_{\theta}\log \pi_{\theta}(m|x_{s})\mathcal{R}_{s}] & \text{sender's gradient} \\
        \nabla_{\phi}\mathcal{L}_{\phi} &= - \mathbb{E}_{\tau}[\nabla_{\phi}\log \rho_{\phi}(a|m,x_{r})\mathcal{R}_{r}] & \text{receiver's gradient}
    \end{array}
\right.
    \end{align}
    In practice, the quantities $(1)$ $l_{\theta}=\log \pi_{\theta}(m|x_{s})|\mathcal{R}_{s}|_{SG}$ and $(2)$ $l_{\phi}=\log \rho_{\phi}(a|m,x_{r})|\mathcal{R}_{r}|_{SG}$ are computed over a batch of game episodes and passed to each agent optimizer. $|.|_{SG}$ is the stop gradient operator that prevents an optimizer from computing the gradient inside the operator.

\paragraph{Optimizing the losses} The optimization encounters challenges, for which we provide a few recipes to ensure a successful optimization process:

\begin{figure}[ht!]
    \centering
    \includegraphics[width=1.\textwidth]{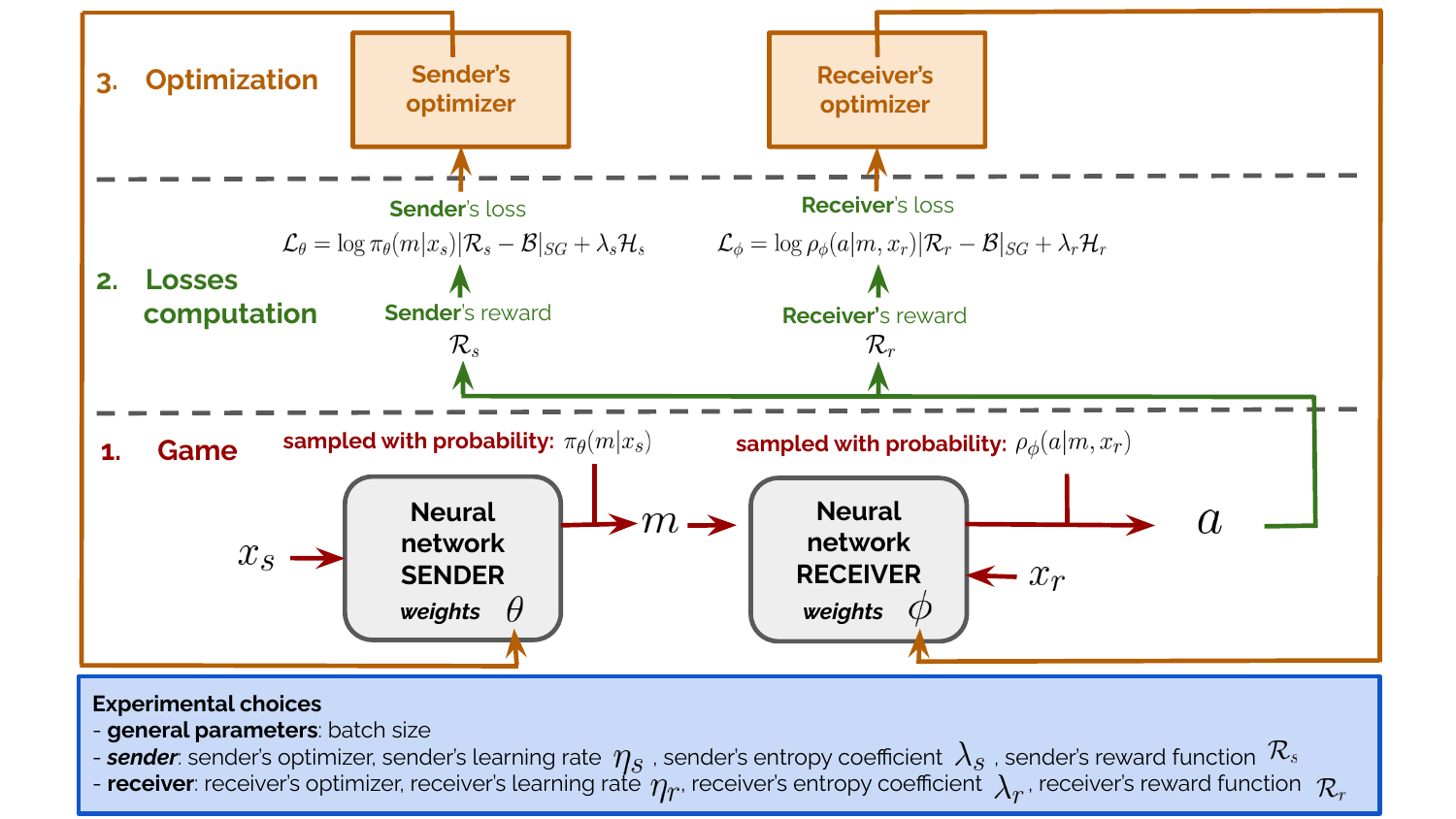}
    \caption{Scheme of the training loop of a communication game optimized with reinforcement learning. \textbf{Step 1 (game):} The sender and receiver perform an episode of the game on a batch of data. Agent's observations $x_{s}$ and $x_{r}$, receiver's action $a$, the probabilities $\pi_{\theta}(m|x_{s})$ and $\rho_{\phi}(a|m,x_{r})$ with which the sender and receiver respectively samples the message $m$ and performs action $a$, the entropy of the sender's policy $\mathcal{H}_{s}$ and receiver's policy $\mathcal{H}_{r}$ are kept to compute the losses. 
    \textbf{Step 2 (losses computation)}: Based on those quantities, we compute sender's reward $\mathcal{R}_{s}$ and receiver's reward $\mathcal{R}_{r}$ and then sender's loss $\mathcal{L}_{\theta}$ and receiver's loss $\mathcal{L}_{\phi}$.
    \textbf{Step 3 (optimization)}: We pass the losses to agents' optimizers that update the weights of the two agents.
    All the experimental choices are framed in blue.}
    \label{fig:optim_game}
\end{figure}

\begin{enumerate}
    \item \textbf{Implementing Policy Gradient} While RL notations may become overwhelming for beginners, their implementation is quite straightforward in practice with recent machine learning libraries~\citep{paszke2019pytorch,jax2018github}.
    \item \textbf{Dealing with large variance} Estimating the gradient of a RL loss is difficult due to the large variance of gradient estimates. Large batch sizes and the baseline method should be used to alleviate this. The latter implies subtracting a baseline $\mathcal{B}$ from the reward $\mathcal{R}$, which does not bias the estimate while reducing the variance. A common baseline is the average value of the reward across a batch of data.
    \item \textbf{Controlling the exploration-exploitation trade-off} To prevent the collapse of training due to sub-optimal average reward, one can control the exploitation-exploration trade-off by penalizing the entropy of the policies with the terms $\lambda_{s}\mathcal{H}_{s}$ and  $\lambda_{r}\mathcal{H}_{r}$  ($\mathcal{H}_{s}$ and $\mathcal{H}_{r}$ refers to the entropy function applied on agents' policies)\footnotemark{}{}. By increasing the coefficient $\lambda_{s}$ (resp. $\lambda_{r}$), the sender's policy (resp. receiver's policy) is encouraged to explore multiple actions instead of focusing on single ones.
\end{enumerate}

\footnotetext{$\mathcal{H}_{s}=\mathcal{H}(\pi_{\theta}(.|x))$ and $\mathcal{H}_{r}=\mathcal{H}(\rho_{\phi}(.|m,x_{r}))$ where $\mathcal{H}$ is the entropy function}

As summarized in Figure~\ref{fig:optim_pip}, the following optimization protocol can be built applying those practices: 

\begin{itemize}
    \item Choose a batch size and for each agent: learning rates $\eta_{s}$ and $\eta_{r}$, reward functions $\mathcal{R}_{s}$ and $\mathcal{R}_{r}$, exploration coefficients $\lambda_{s}$ and $\lambda_{r}$;
    \item Iteratively:
    \begin{enumerate}
    \item Perform a game episode on a batch of data;
    \item Compute the losses: 
    \begin{align*}
        \left\{
    \begin{array}{lll}
        \log \pi_{\theta}(m|x_{s})|\mathcal{R}_{s}-\mathcal{B}|_{SG} + \lambda_{s}\mathcal{H}_{s} & \text{Sender's loss} \\
        \log \rho_{\phi}(a|m,x_{r})|\mathcal{R}_{r}-\mathcal{B}|_{SG} + \lambda_{r}\mathcal{H}_{r} & \text{Receiver's loss} 
    \end{array}
\right.
    \end{align*}
    \item Pass sender's loss (resp. receiver's loss) to sender's optimizer (resp. receiver's optimizer), which performs a parameters update for each agent.
\end{enumerate}
\end{itemize}

\remark{Training a communication game involves selecting reward functions for each agent and tuning numerous parameters: the batch size, agents' learning rate, and exploration coefficient. The initial step in a simulation is to identify a set of parameters that allows the agents to solve the task. However, it is essential to consider how these choices affect the system's overall training dynamics. \cite{rita2022on,rita2022emergent} demonstrate that optimization decisions, especially asymmetries between the sender and the receiver, crucially impact the properties of the emergent communication protocol.
}

\section{Case study: Simulating a Visual Discrimination Game}
\label{sec:case_study}

\begin{figure}[ht!]
    \centering
    \includegraphics[width=1.\textwidth]{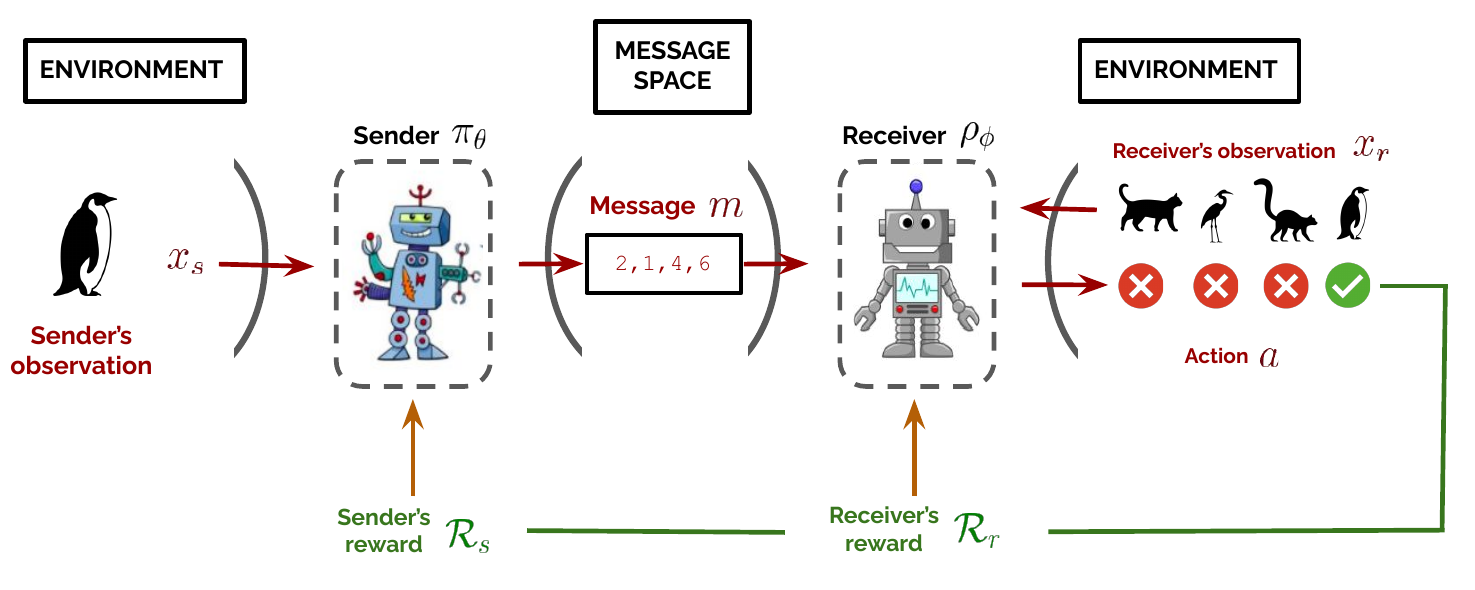}
    \caption{In the Visual Discrimination Game, the Sender sees an image (sender observation) and communicates about it to the receiver. Using the message, the receiver has to guess the target image among a set of $N$ candidate images (receiver observation). If the guess (receiver action) is identical to the sender's image, the two agents are equally rewarded}
    \label{fig:case_study}
\end{figure}

We now focus on a particular communication game: the Visual Discrimination Game, a type of Lewis Referential Games~\citep{lewis1969convention}. These games, which explore how languages emerge through their use, have been extensively studied from theoretical and experimental angles in language evolution~\citep{crawford1982strategic,blume1998experimental,skyrms2010signals,raviv2019larger}.

\paragraph{Game rules} 
The Visual Discrimination Game involves two players: a sender and a receiver. The game proceeds as follows:
\begin{itemize}
    \item The sender sees an image and communicates about it to the receiver;
    \item Using the message, the receiver has to guess the original image seen by the sender among a set of $N$ candidate images;
    \item The original image is revealed, and the two players are informed about the task's success.
\end{itemize}
Agents play the game repeatedly until they synchronize on a communication protocol that enables the receiver to distinguish any image from any set of distractors.

\paragraph{Designing the game} The following parameters must be specified: 
\begin{itemize}
    \item \textbf{Image dataset} This is the set of images the agents must communicate about. Compared to human simulations, machine learning experiments can be conducted with large-scale datasets compared to human simulations, which is critical for developing a rich communication protocol. For example, some studies, such as \cite{lazaridou2016multi,dessi2021interpretable,chaabouni2022emergent,rita2022emergent}, have relied on ImageNet~\citep{deng2009imagenet,russakovsky2015imagenet} ($14$ million images dataset spanning more than $20,000$ categories including animals, vehicles, objects or instruments). Synthetic datasets, like CLEVR~\citep{johnson2017clevr} are also valuable for evaluating agents' ability to communicate about ambiguous images using compositional languages.
    \item \textbf{Number of candidate images $N$} The receiver must differentiate the original image from $N-1$ distractor images. The task's difficulty depends on the value of $N$: a higher $N$ requires a more precise communication protocol.  
    \item \textbf{Message space} The message space is shaped by the vocabulary $V$ and message maximum length $L$. Adjusting those parameters crucially influences the sender's expressiveness. By denoting the vocabulary size by $|V|$, the sender can use a total number of  $L^{|V|}$ messages.
\end{itemize}

\begin{figure}[ht!]
    \centering
    \includegraphics[width=1.\textwidth]{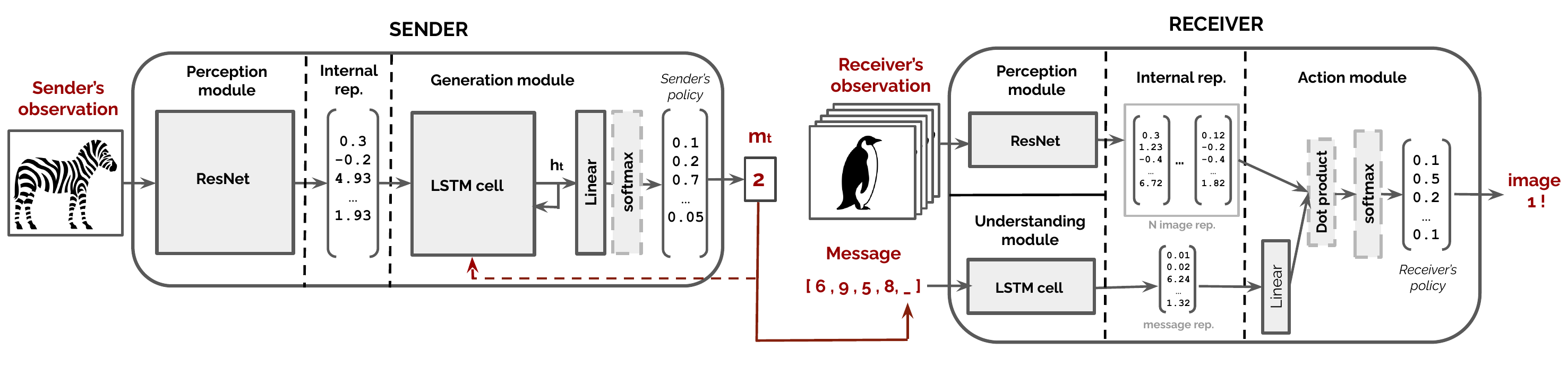}
    \caption{Examples of neural communicative agents in the Visual Discrimination Game. Module choices are based on the usual agent design from the literature. 
    \vspace{0.5em}
    \\
    (left) The sender uses a perception module and a generation module with an internal representation in between. The perception module is a ResNet~\citep{he2016deep}, i.e., CNN with additional skip connections, which produces image representations. The generation module consists of a one-layer LSTM~\citep{hochreiter1997long}, an RNN variant, followed by a linear layer and a softmax, which generates the sender's policy (probability distribution over the vocabulary symbols). The output is obtained by recursively sampling symbols until the maximum length is reached or the end-of-sentence token is generated. 
    \vspace{0.5em}
    \\ 
    (right) The receiver comprises a perception module, an understanding module, and an action module with internal representations in between. Like the sender, the perception module is a ResNet, and the understanding module is a one-layer LSTM. The action module includes a linear layer that maps the message representation to a vector with the same dimension as image representations. This dot product compares each image representation to the message representation. Finally, a softmax transforms the resulting $N$ values to probability distributions over the $N$ possible actions. The action, i.e., image choice, is obtained by sampling according to this distribution.}
    \label{fig:agents_game}
\end{figure}

\paragraph{Game formalism} Using previous notations:
\begin{itemize}
    \item \textit{Sender's observation} $x_{s}$ is an image sampled from the dataset.
    \item \textit{Receiver's observation} $x_{r}$ is a set of $N$ images sampled from the dataset that includes sender's observation $x_{s}$.
    \item \textit{Message} $m$ is a message sent by the sender.
    \item \textit{Action} $a$ is the choice of image among the set of $N$ images.
\end{itemize}

\paragraph{Designing the network} Following agents design of Figure~\ref{fig:summary_NN} and neural networks descriptions in Figure~\ref{fig:NNs}, Figure~\ref{fig:agents_game} reports standard agents design choices in the Visual Discrimination Game.

\goodpractice{\textit{Scaling-up neural networks to natural inputs}
\begin{itemize}
\item \textbf{Pretrained Network}: Pre-trained modules are employed to simplify computation. These are neural networks that have been previously trained on a different task. In our example, a pre-trained network is used as a perception module, which is frozen during the training~\citep{chaabouni2022emergent,rita2022emergent}. It aims to provide meaningful image representations inherited from the previous task. \citet{chaabouni2022emergent} provide such pre-trained representations for the ImageNet~\citep{deng2009imagenet,russakovsky2015imagenet} and CelebA~\citep{liu2015deep} datasets.
\item  \textbf{Multimodal Learning:} Fusing different inputs and outputs is referred to as Multimodal Learning~\citep{baltruvsaitis2018multimodal}. For instance, the receiver needs message and image representations to perform an action. Simple methods like concatenation or scalar product are often used to fuse different neural module representations. When scaling up to more realistic tasks, more advanced fusing mechanisms are sometimes required, e.g., modulation~\citep{dumoulin2018feature} or multi-modal transformers~\citep{lu2019vilbert,alayrac2022flamingo}.
\end{itemize}
}


\paragraph{Optimization} Using Figure~\ref{fig:optim_game} scheme, a working optimization algorithm using reinforcement learning only is described in Algorithm~\ref{alg:cap}.

\begin{algorithm}
\caption{Visual Discrimination Game optimization}\label{alg:cap}
\begin{algorithmic}
\Require $\pi_{\theta}$ (sender), $\rho_{\phi}$ (receiver) \Comment{Trainable neural networks}
\Require $\mathcal{D}_{train}$ (image dataset), batch\_size, $N$, $V$, $L$ \Comment{General parameters}
\Require sender's optimizer, $\eta_{s}$ , $\mathcal{R}$, $\lambda_{s}$, \Comment{Sender's parameters}
\Require receiver's optimizer, $\eta_{r}$ , $\mathcal{R}$, $\lambda_{r}$, \Comment{Receiver's parameters}
\\
\While{convergence} \Comment{\textbf{Learning loop}}
    \State \textbf{1. Game}
    \State $x_{s} \sim \mathcal{D}_{train}$, $x_{r} \sim \mathcal{D}_{train}$ \Comment{Images sampled from the dataset}
    \State $\tau \sim \textrm{game}(\pi_{\theta},\rho_{\phi},x_{s},x_{r},V,L)$ \Comment{Game episode (batch\_size games in parallel)}
    \\ 
    \State \textbf{2. Losses computation}
    \State $\mathcal{R} \gets f_{reward}(\tau)$ \Comment{Reward computation}
    \State $\mathcal{L}_{s} \gets \log \pi_{\theta}(m|x_{s})|\mathcal{R}-\mathcal{B}|_{SG} + \lambda_{s}\mathcal{H}_{s}$ \Comment{Sender's loss}
    \State $\mathcal{L}_{r} \gets \log \rho_{\phi}(a|m,x_{r})|\mathcal{R}-\mathcal{B}|_{SG} + \lambda_{r}\mathcal{H}_{r}$ \Comment{Receiver's loss}
    
    \\
    \State \textbf{3. Optimizer}
    \State $\theta \gets \text{sender's optimizer}(\theta, \eta_{s},\mathcal{L}_{s})$ \Comment{Sender's weights update}
    \State $\phi \gets \text{receiver's optimizer}(\theta, \eta_{r},\mathcal{L}_{r})$ \Comment{Receiver's weights update}
\EndWhile
\end{algorithmic}
\end{algorithm}

\paragraph{Parameter choices} A typical reward function assigns a reward of $1$ if the receiver picks up the correct image and $0$ otherwise. The modeling parameters, which include the vocabulary $V$, maximum message length $L$, and the number of candidates $N$, should be selected based on the problem under investigation.
For the optimization, we recommend using a large batch size ($512$ or $1024$ typically) and one Adam~\citep{kingma2014adam} optimizer per agent. 
The other parameters, including exploration coefficients $\lambda_{s}$, $\lambda_{r}$ and learning rates $\eta_{s}$, $\eta_{r}$ are interdependent and should be adjusted simultaneously until the simulation works. Common strategies for parameter tuning include manual adjustment or more systematic methods like grid search~\citep{feurer2019hyperparameter}.

\paragraph{Implementation} A full implementation of the game with technical details and a starting set of working parameters is provided at:
{\begin{center} \url{https://github.com/MathieuRita/LangageEvolution_with_DeepLearning}\end{center}}

\section{Bridging the gap between neural networks and humans in language evolution simulations}
\label{sec:research}

This section focuses on current endeavors in using deep learning as a framework for language evolution simulations. 
It covers the field's progress in using neural networks to replicate human languages and highlights the potential and challenges of deep learning simulations.

\subsection{Opportunities opened by deep learning simulations}

\paragraph{The control/realism duality of simulations} Neural network simulations provide extensive flexibility for modeling various aspects of language emergence simulations, including the game, inputs, and agents. Two primary strategies have been pursued: simplifying experiments into controllable settings~\citep{kottur2017natural,chaabouni2019anti,chaabouni2020compositionality,ren2020compositional,rita2022on}, assessing the influence of incremental modeling elements; and creating more humanly plausible scenarios that emulate language emergence in complex environments~\citep{das2019tarmac,jaques2019social}. It has resulted in various tasks, from basic referential tasks to complex ecological tasks in grounded environments~\citep{das2019tarmac}. In terms of inputs, it spans from hand-designed structured and controllable inputs~\citep{kottur2017natural,chaabouni2019anti,chaabouni2020compositionality,ren2020compositional,rita2020lazimpa,rita2022on} to complicated visual inputs~\citep{evtimova2017emergent,lazaridou2018emergence,dessi2021interpretable,chaabouni2022emergent,rita2022emergent}. As for agents, it extends from pairs of agents decomposed into senders and receivers to pairs of bidirectional agents~\citep{bouchacourt-baroni-2018-agents,graesser2019emergent,taillandierneural,michelrevisiting} and populations~\citep{tieleman2019shaping,graesser2019emergent,rita2022on,michelrevisiting}.

\paragraph{Evaluating emergent linguistic phenomena} Simulations give rise to the emergence of artificial languages whose properties are compared to human languages.
As human languages can be described in terms of language universals, i.e., abstract properties found across all human languages, studies have tried to establish the conditions under which those universal properties emerge. Such universals mainly include compositionality, i.e., the ability to decompose the meaning of
an utterance as a function of its constituents~\citep{hockett1960origin}, measured through topographic similarity~\citep{brighton2006understanding}, ~\citep{chaabouni2020compositionality}, or Tree Reconstruction Error~\citep{andreas2019measuring}; efficiency, i.e., efficient information compression, measured through message length statistics and semantic categorization \citep{zipf1949principie, regier2015word}; demographic trends, such as the impact of population size, contact agents proportion, network topology on language structure~\citep{clyne1992linguistic,wray2007543,wagner2009communication,Lupyan2010}.

\subsection{Do neural networks replicate human behaviors?}
\label{subsec:replication}

To provide valuable insights through deep learning simulations, replicating human languages is essential. This involves identifying the basic assumptions needed for artificial agents to display human-like language patterns in their communication protocols.

\paragraph{The referential objective is insufficient for the emergence of natural language features} 
A first approach is to question whether the most simple communication task, i.e., referring to objects in an environment through referential communication, is enough to see human language features emerge. 
The first works on referential tasks showed that neural agents could successfully derive a communication protocol from solving the task~\citep{kottur2017natural,lazaridou2016multi,havrylov2017emergence}. Still, such protocols are neither interpretable nor bear the core properties of human languages.
Indeed, agents tasked with communicating about images do not utilize semantically significant concepts but instead shortcut the task by basing their messages on low-level visual features~\citep{lazaridou2016multi,havrylov2017emergence,chaabouni2022emergent,bouchacourt-baroni-2018-agents}. 
%
Additionally, when agents communicate about hand-designed structured sets of objects in a simple referential task, fundamental properties of natural languages such as compositionality~\citep{kottur2017natural,chaabouni2020compositionality} or efficiency~\citep{zipf2016human} do not spontaneously arise~\citep{chaabouni2019anti}. 
Eventually, when referential games are played within a population of agents, human demographic trends are not reproduced. Population size does not behave as a regularization factor~\citep{li2019ease,cogswell2019emergence,rita2022on,chaabouni2022emergent} and agents do not synchronize on a shared protocol~\citep{rita2022on,michelrevisiting}. Understanding the origins of these discrepancies from either an optimization or modelization perspective is an active research question.

\paragraph{Incorporating human-inspired constraints drive the emergence of natural language features}

To recover human languages features, different human-inspired constraints have incrementally been added to simulations.
Inspired by Iterated Learning~\citep{kirby2001spontaneous,kirby2014iterated}, one line of research has explored the effects of learnability constraints on language emergence by altering learning dynamics.
\cite{li2019ease,ren2020compositional,cogswell2019emergence} implement neural variants of Iterated Learning by periodically introducing newborn agents and mimicking generational transmission. They find that those learning constraints drive the selection of more compositional languages, as they are easier to learn~\citep{li2019ease}.
%
Another line of research focuses on incorporating cognitively inspired biases into agent modeling. For example, \cite{rita2020lazimpa} show that Zipf's Law of Abbreviation~\citep{zipf1949principie} emerges when both pressures toward Least Effort production~\citep{zipf1949principie,piantadosi2011word,kanwal2017zipf} and comprehension laziness are introduced. 
Eventually, some researchers have refined population modeling. \cite{rita2022on} introduce learning speed variations into populations and recover the relationship between population size and language structure reported in previous works~\citep{Lupyan2010,meir2012influence,WALS,Reali2018,raviv2019larger}. \cite{graesser2019emergent} examine contact agents phenomena and show that a contact language can either converge towards the majority protocol or result in novel creole languages, depending on the inter- and intra-community densities. 
\cite{kim2021emergent} and \cite{michelrevisiting} study how social graph connectivity impacts the development of shared languages.

\subsection{Toward realistic experiments}

Although incorporating human-inspired constraints shows promise for replicating human language features, the simplicity of current models remains limited. An avenue is opened for the design of humanly plausible experiments. We present efforts to build more realistic models and discuss the associated challenges here.



\paragraph{Toward realistic scenarios} Task-specific communication games may be restrictive as they overlook other aspects of our language, such as conversation, interaction with the physical world, and other modalities. More realistic scenarios are needed to encompass all aspects of our language.
Some attempts have been made to create more plausible settings. \cite{chaabouni2022emergent} complexify the referential task by scaling the game to large datasets and tasking agents to retrieve images among $1000$ distractors.
\cite{evtimova2017emergent,taillandierneural} model conversation by building bidirectional agents for multi-turn communications; \cite{bullard2020exploring} explore nonverbal communication using spatially articulated agents; \citep{das2019tarmac} ground agents in more realistic 2D and 3D environments; \cite{jaques2019social} test agents ability to solve social dilemmas in grounded environments. 
However, making more realistic games poses both technical and analytical challenges. Training instabilities can occur when games become more complex, requiring optimization tricks~\citep{chaabouni2022emergent}. Moreover, as environments become more complex, the emergence of language is more challenging to analyze. For example, \cite{lowe2020interaction} demonstrates how agents can solve complex tasks with shallow communication protocols and why new tools are needed to assess emergent languages qualitatively and quantitatively in these situations.

\paragraph{Toward realistic agents}
Many neural communication agents are designed for specific games and lack crucial aspects of human cognition. For instance, agents are often limited to either speaking or listening, which overlooks the interplay between comprehension and production~\citep{galke2022}. 
Some works propose more realistic agents.
These include bidirectional agents that both speak and listen~\citep{bouchacourt-baroni-2018-agents,graesser2019emergent,michelrevisiting,taillandierneural}, as well as agents with restricted memory capacity that better mirrors human cognition~\citep{resnick2019capacity}. Additionally, \cite{rita2020lazimpa} incorporate the Least Effort Principle to make agents efficient encoders~\citep{zipf1949principie,piantadosi2011word,kanwal2017zipf}.
Still, despite the impact of these modeling constraints on emergent language properties, they are not consistently applied across the literature.

\paragraph{Toward linguistically informed metrics} One of the main limitations of neural emergent languages is that current metrics may not capture crucial features of human languages. For instance, most work only uses topographic similarity~\citep{lazaridou2018emergence, li2019ease} as a structural metric~\citep{brighton2006understanding}, which assumes that the units of the message carry out the meaning. In human languages, the meaning units are the results of a combinatorial process using nonmeaningful units, such as phonetic features or phonemes (the so-called double articulation phenomenon~\citep{martinet1960elements}; or duality of patterning~\citep{hockett1970leonard}). Other universal properties of language (formal universals~\citep{chomsky1968sound}) include the reliance on symbols and rules~\citep{fodor1988connectionism}, the use of hierarchical representations or long distance dependencies~\citep{hauser2002faculty}, the existence of part-of-speech classes~\citep{rijkhoff2007word} such as the distinction between content and grammatical words, the existence of deixis~\citep{lyons1977semantics}, i.e. the use of certain parts of the message to refer to places or time or person relative to the context of elocution of the message, and many others.
Studying such properties is challenging as it requires the design of adapted measures that could be computed both on human and artificial languages. Furthermore, current artificial settings are often too simple to drive the emergence of such properties, reinforcing the need for more realistic scenarios that translate into our environment's complexity.

\paragraph{Bridging Natural Language Processing and Language Emergence}

One area of research focuses on investigating whether language emergence simulations can potentially enhance natural language processing tasks. One approach involves pre-training language models with artificial languages that emerged from communication games, resulting in a moderate boost when fine-tuning low-resource language tasks~\citep{yao2022linking}. Another approach is exploring machine-machine interaction to learn an emergent communication protocol that prompts large language models~\citep{shin2020autoprompt,deng2022rlprompt}. Reciprocally, natural language models can be utilized to explore language evolution from pre-trained languages, such as studying creolization~\citep{armstrong2022jampatoisnli} or language drift phenomena~\citep{lu2020countering}. Finally, at the time of writing, Large Language Models (LLM)~\citep{ouyang2022training,hoffmann2022training,touvron2023llama,bai2022constitutional} have demonstrated potential for Natural Language P rocessing, as they can handle multiple languages and perform basic reasoning. This presents exciting opportunities for language emergence research from scientific and practical perspectives \citep{baroni2022emergent}.

\section{Conclusion}
Deep learning advancements offer new opportunities for simulating language evolution, as neural networks can handle diverse data without pre-defined human priors. They scale significantly regarding dataset size, task complexity, and number of participants or generations. This opens up possibilities for creating realistic language evolution scenarios at unprecedented scales.
Reciprocally, language evolution research can provide valuable insights for developing future deep learning models. In the journey toward building intelligent language models, it seems essential to incorporate constraints and mechanisms that shape the development and evolution of language, such as perceptual, social, or environmental pressures. 
We hope this chapter will encourage researchers in both language evolution and deep learning to collaborate and jointly explore those two captivating black-boxes: humans and neural networks.

\bibliography{biblio}

\end{document}